\documentclass[manuscript,screen]{acmart} 
\usepackage{algorithm}
\usepackage{algpseudocode}
\usepackage[inline]{enumitem}
\graphicspath{{image/}}
\usepackage{csvsimple}
\usepackage{subfig}
\usepackage{graphicx}
\usepackage{multicol}
\usepackage{multirow}
\usepackage{graphicx}
\usepackage{tikz}
\usepackage{forest}
\usetikzlibrary{trees,positioning,shapes,shadows,arrows.meta}
\usepackage{smartdiagram}
\usepackage{xcolor}
\colorlet{myblue}{blue!70!black}
\colorlet{mylightblue}{blue!10}
\colorlet{branch}{green!30!black}
\colorlet{evolcol}{green!50!black}
\colorlet{natalcol}{red!50!white!60!black}
\colorlet{hormcol}{orange!70!black}
\colorlet{ethcol}{yellow!80!black}
\colorlet{stimcol}{red!80!black}
\colorlet{neurcol}{blue!80!black}
\AtBeginDocument{%
  \providecommand\BibTeX{{%
    \normalfont B\kern-0.5em{\scshape i\kern-0.25em b}\kern-0.8em\TeX}}}

\setcopyright{acmcopyright}
\copyrightyear{2018}
\acmYear{2018}
\acmDOI{XXXXXXX.XXXXXXX}

\acmConference[Conference acronym 'XX]{Make sure to enter the correct
  conference title from your rights confirmation emai}{}{}
\acmPrice{15.00}
\acmISBN{978-1-4503-XXXX-X/18/06}

\begin{document}

\title{Survey on Leveraging Uncertainty Estimation Towards Trustworthy Deep Neural Networks: The Case of Reject Option and Post-training Processing}

\author{Mehedi Hasan}
\email{mmhasan@deakin.edu.au}
\orcid{0000-0001-7721-0258}
\affiliation{%
  \institution{Institute for Intelligent Systems Research and Innovation (IISRI), Deakin University}
  \city{Geelong}
  \state{Victoria}
  \country{Australia}
  \postcode{3220}
}
\author{Moloud Abdar}
\email{m.abdar1987@gmail.com \& m.abdar@deakin.edu.au}
\orcid{0000-0002-3059-6357}
\affiliation{%
  \institution{Institute for Intelligent Systems Research and Innovation (IISRI), Deakin University}
  \city{Geelong}
  \state{Victoria}
  \country{Australia}
  \postcode{3220}
}

\author{Abbas Khosravi}
\email{abbas.khosravi@deakin.edu.au}
\orcid{0000-0001-6927-0744}
\affiliation{%
  \institution{Institute for Intelligent Systems Research and Innovation (IISRI), Deakin University}
  \city{Geelong}
  \state{Victoria}
  \country{Australia}
  \postcode{3220}
}

\author{Uwe Aickelin}
\email{uwe.aickelin@unimelb.edu.au}
\orcid{0000-0002-2679-2275}
\affiliation{%
  \institution{School of Computing and Information Systems, The University of Melbourne}
  \city{Melbourne}
  \state{Victoria}
  \country{Australia}
  \postcode{3010}
}

\author{Pietro Liò}
\email{pl219@cam.ac.uk}
\orcid{0000-0002-0540-5053}
\affiliation{%
  \institution{Department of Computer Science and Technology, University of Cambridge}
  \city{Cambridge}
  \state{Cambridge}
  \country{United Kingdom}
  \postcode{CB3 0FD}
}

\author{Ibrahim Hossain}
\email{i.hossain@deakin.edu.au}
\affiliation{%
  \institution{Institute for Intelligent Systems Research and Innovation (IISRI), Deakin University}
  \city{Geelong}
  \state{Victoria}
  \country{Australia}
  \postcode{3220}
}

\author{Ashikur Rahman}
\affiliation{%
  \institution{Department of Computer Science and Engineering, Bangladesh University of Engineering and Technology}
  \city{Dhaka 1000}
  \country{Bangladesh}}
\email{ashikur@cse.buet.ac.bd}
\orcid{0000-0003-4231-6065}

\author{Saeid Nahavandi}
\email{saeid.nahavandi@deakin.edu.au}
\orcid{0000-0002-0360-5270}
\affiliation{%
  \institution{Institute for Intelligent Systems Research and Innovation (IISRI), Deakin University}
  \city{Geelong}
  \state{Victoria}
  \country{Australia}
  \postcode{3220}
}

\renewcommand{\shortauthors}{Hasan et al.}

\begin{abstract}
Although neural networks (especially deep neural networks) have achieved \textit{better-than-human} performance in many fields, their real-world deployment is still questionable due to the lack of awareness about the limitation in their knowledge. To incorporate such awareness in the machine learning model, prediction with reject option (also known as selective classification or classification with abstention) has been proposed in literature. In this paper, we present a systematic review of the prediction with the reject option in the context of various neural networks. To the best of our knowledge, this is the first study focusing on this aspect of neural networks. Moreover, we discuss different novel loss functions related to the reject option and post-training processing (if any) of network output for generating suitable measurements for knowledge awareness of the model. Finally, we address the application of the rejection option in reducing the prediction time for the real-time problems and present a comprehensive summary of the techniques related to the reject option in the context of extensive variety of neural networks. Our code is available on GitHub: \url{https://github.com/MehediHasanTutul/Reject_option}.
\end{abstract}




\maketitle

\section{Introduction}
The field of machine learning, especially Neural Networks (NNs), has been revolutionized in the last decade due to abundance of data and computing power \cite{Goodfellow-et-al-2016}. One of the main goals of NNs research has been to improve the \textit{performance} of the model, such as prediction accuracy. However, estimating some form of scores denoting the \textit{quality} of the prediction is also of paramount significance for many real world applications such as medical image analysis. When a model is aware of the quality of its prediction, it can decide whether to predict or refrain from prediction based on the quality factors as incorrect prediction can be much costly compared to no prediction \cite{abdar2021uncertainty, abdar2021barf, abdar2022need}. It is similar to the situation when a medical expert is shown an x-ray image of a patient but the expert can not make any conclusive decision due to lack of sufficient information or some other issues.

\begin{figure}[!t]
\centering
\includegraphics[scale=0.55]{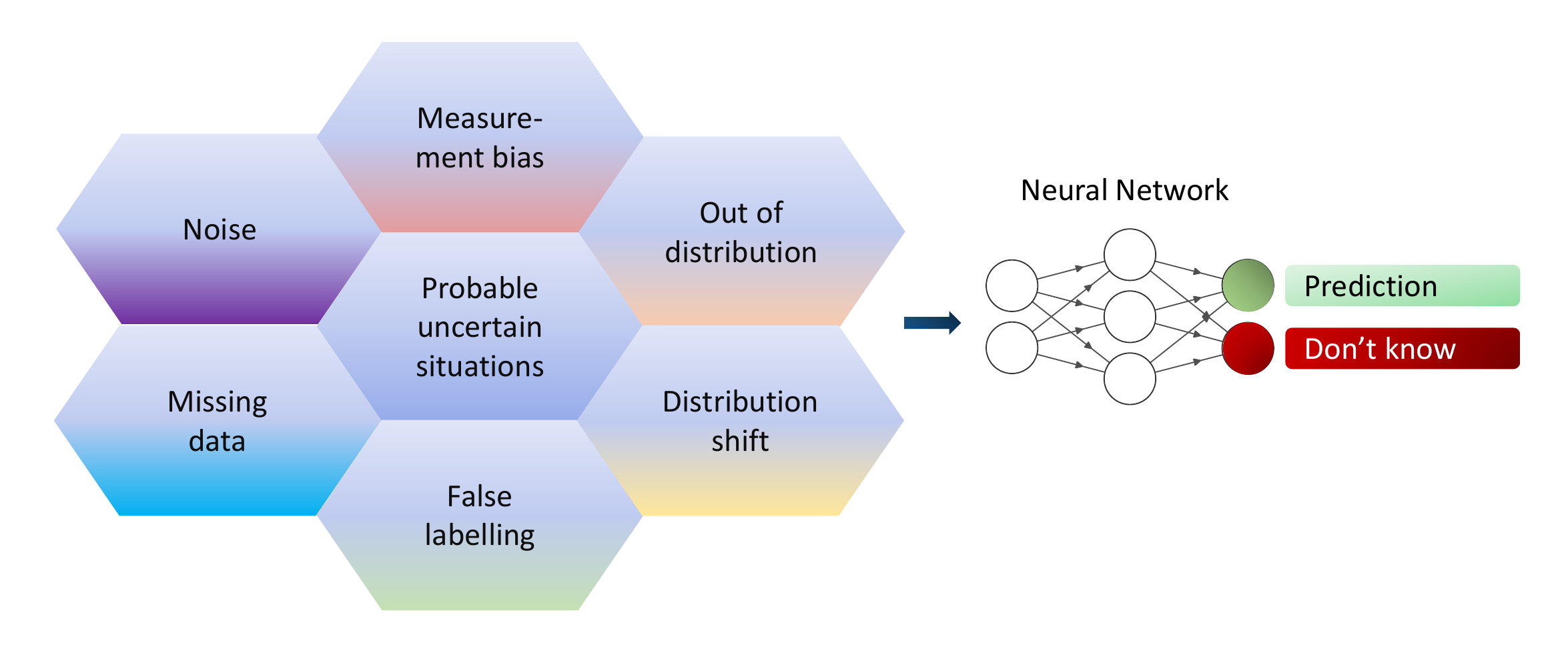}
\caption{An overview of reliable Neural Network. If it is fed on an input which may incur incorrect prediction, it should say- "I don't know" rather than predicting incorrectly.}
\label{reliable_NN}
\end{figure}

Similarly, when a machine learning model (such as a DNN) is deployed in real-world, we typically cannot control inputs fed to it  during its operation. Hence, one may want the model to behave similar to the expert in the example described above. The expected behavior of such a DNN/NN in uncertain situations is illustrated in Fig. \ref{reliable_NN}.
The subfield in machine learning that deals with incorporating this capability is known as prediction with a \textit{reject option} (or selective prediction or abstention prediction). One of the applications of the reject option is fraud detection in financial transactions, where the abnormal transactions can be identified without further processing. Other possible applications include increasing the reliability of face recognition, handwriting, and speech recognition systems.

There are many different approaches to address the issue of robustness of modern DNNs in the presence of noisy real-world data. It is demonstrated that he effective capacity of DNNs is sufficient to fit random noise without increasing the training time substantially compared to real data. Furthermore, DNNs can also memorize the entire real dataset \cite{zhang2021understanding}. Later, further investigation into this area reveals that DNNs learns useful patterns for real dataset before fitting to noise \cite{ pmlr-v70-arpit17a}. Leveraging this property of DNNs, an effective and end-to-end target modification approach ProselfLC \cite{ DBLP:journals/corr/abs-2005-03788, 10.48550/arxiv.2207.00118} is developed. It has also been shown that the example weighting is a built-in feature in empirical loss functions, i.e., Mean Absolute Error (MAE) shows noise robustness by emphasizing on uncertain example more than certain samples \cite{ DBLP:journals/corr/abs-1903-12141}. The large-scale real-world dataset being imbalanced and composed of noisy labels, to facilitate training robust DNNs using various example weighting schemes, derivative manipulation is proposed which directly works on gradients bypassing a loss function \cite{DBLP:journals/corr/abs-1905-11233}. All these approaches need separate discussions of their own which is beyond the scope of this work. Though the literature on the prediction with a reject option is quite extensive and has been around since the 1960s \cite{Chow1957AnOC}, it has rarely been studied in the context of NNs. Here, we should clarify that researchers have recently been paying close attention to slightly different sub-fields that also deal with the issue of acceptability of the prediction made by a NN such as \textit{uncertainty quantification (UQ) and out-of-distribution (OOD) detection}. There are a notable amount of review works available in both of these sub-fields \cite{10.48550/arxiv.2110.11334,ABDAR2021104418}. 

In this review paper, we cover the reject option or selective classification in the context of NNs. To the best of our knowledge, there has been no review work focusing on this particular aspect of NNs so far. Our goal is that this work can serve as a starting stone for researchers early in their careers in this area. We can summarize our contribution as: \begin{enumerate*}
    \item For the first time, we review the articles related to the reject option or abstention prediction in the context of the NNs;
    \item The rejection or selection strategies are categorized;
    \item We sort out and briefly explain the novel cost functions directly related to the rejection or selection of the output of the NNs;
    \item We summarize different scoring methods and optimization strategies to determine the optimal boundary between the prediction and the rejection region; and
    \item The main research gaps are indicated for future investigation.
\end{enumerate*}

The rest of the paper is arranged as follows: Section \ref{sec:Lit_Search} explains the strategy used to search and sort out related articles. Section \ref{sec:problem_formulation} formulates the rejection problem and categorizes strategies related to the rejection mechanisms. Section \ref{sec:loss_func} discusses novel cost functions involving rejection mechanisms. Then, Section \ref{sec:post_process} focuses on the novel scoring methods and the threshold-choosing strategies for optimal decision-making. Results from comparative numerical study is discussed in \ref{sec:numerical_study}. After that, Section \ref{sec:application} presents some applications of the reject option. Some guidance for future research is provided in Section \ref{sec:future_scope}. Finally, Section \ref{sec:conclusion} concludes the paper.

\section{Literature Search Methodology}
\label{sec:Lit_Search}
In this review, we followed the PRISMA (Preferred Reporting Items for Systematic Reviews and Meta-Analyses) procedure\cite{10.1371/journal.pmed.1000097} to search and scale down the related research works of our interest. To begin, we wish to explain our choice of database. While `Scopus' and `Web of Science' are good databases for comprehensive literature searches, they do not include preprint works, unlike arXiv where many quality unpublished articles are available. Considering this situation, we selected `Engineering Village' database from Elsevier as it includes preprints from arXiv. We completed the search in multiple steps as our target topic is published in in several sub-areas of deep learning. The list of keywords used in our search are:
\begin{itemize}
    \item "selective prediction" OR "selection prediction" OR "selective classification" OR "selection classification" OR reject* classif* OR reject* predict* OR "reject option" OR abstain* OR abstention in title AND ("deep learning" OR network*) in abstract.
    \item (IDK OR "I don't know" OR over-think) in title AND ("neural network" OR "machine learning" OR "deep learning") in abstract.
\end{itemize}

\begin{figure}[!t]
\centering
\subfloat[]{
\includegraphics[scale=0.37]{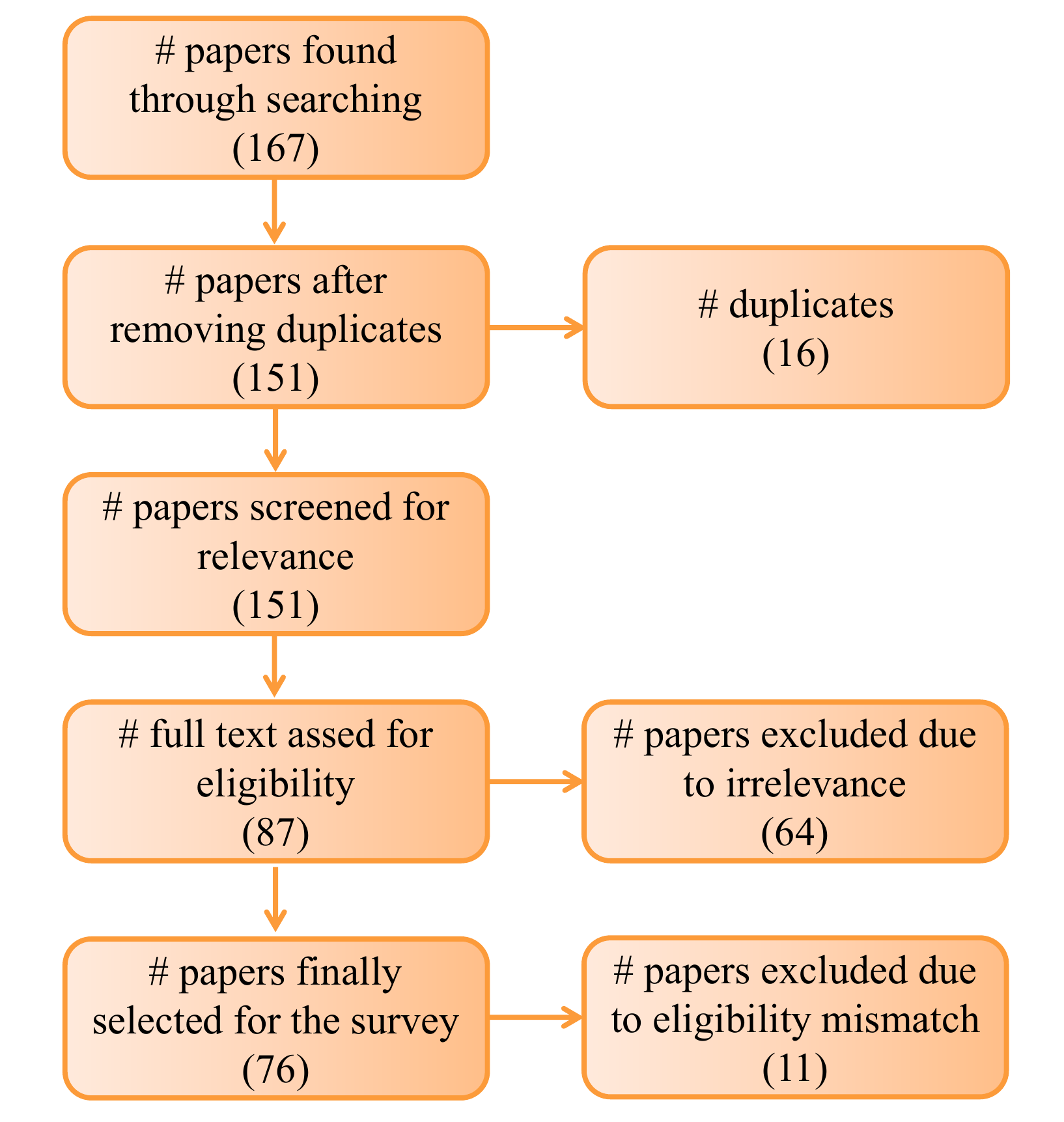}
\label{fig_paper_search}}
\subfloat[]{\includegraphics[scale=0.7]{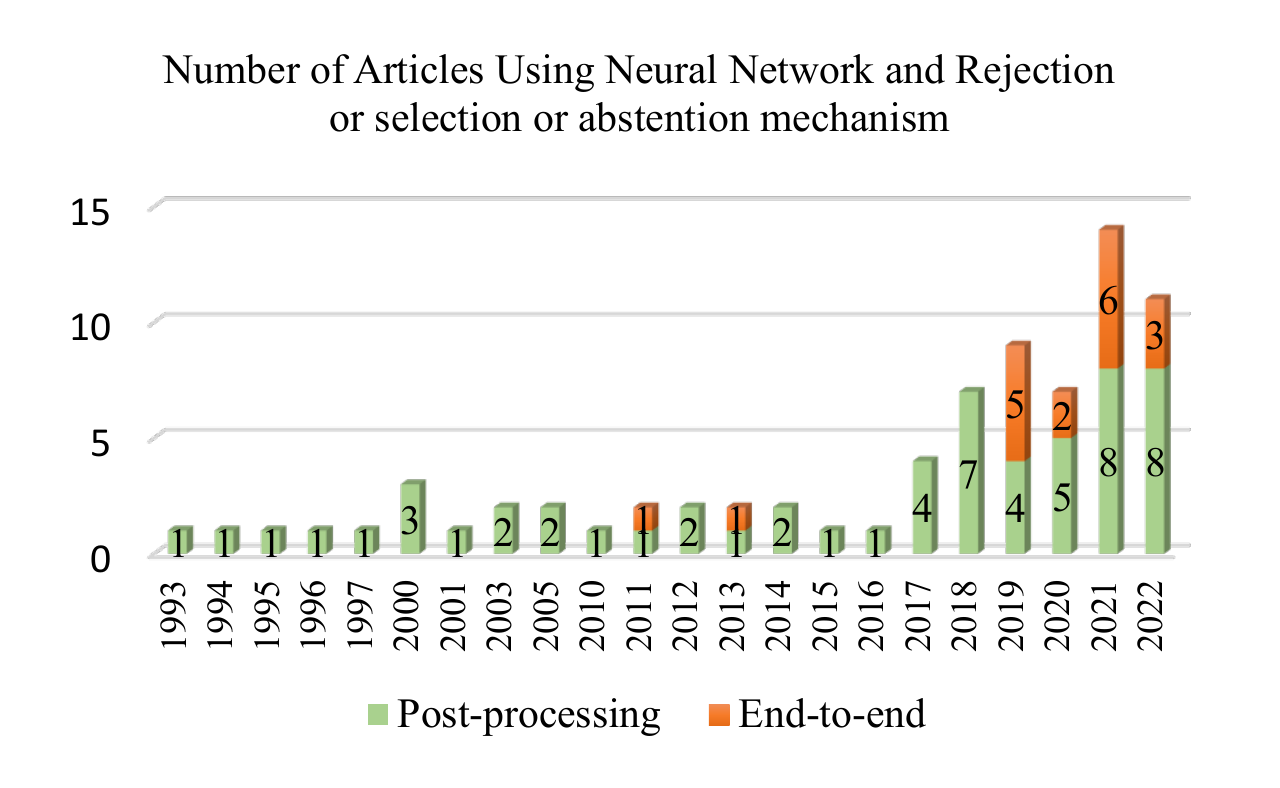}
\label{time_range_paper}}
\caption{(a) Paper selection procedure based of PRISMA process.(b) Number of related papers published in each year. Post-processing: The outputs of the NN are processed to predict with reject option. End-to-end: No processing is required for the outputs.}
\end{figure}

We performed the search on 30 January 2023 without time range restrictions. We removed duplicates, and then going through abstracts, screened the remaining papers for their relevance to the study aims. Finally, we assessed the full text of the selected papers using the following inclusion criteria:
\begin{itemize}
    \item The machine learning model uses an NN.
    \item The final prediction includes a kind of rejection or selection or abstention mechanism on top of the regular network prediction.
\end{itemize}
At the end of this PRISMA procedure, 76 papers were selected for review. It should be noted that there are plenty of works in rejection classification in the machine learning field where SVM and other ML models are used \cite{https://doi.org/10.48550/arxiv.2006.16597,https://doi.org/10.48550/arxiv.2107.11277,NIPS2008_3df1d4b9}. As the main focus of our study is the idea of rejection in the case of NNs, we did not include the other classical machine learning methods in our search. It is clear from Fig. \ref{time_range_paper} that the research in this area has peaked since 2016.


\section{Problem formulation and classification}
\label{sec:problem_formulation}
\begin{figure}[!t]
\centering
\includegraphics[scale=0.35]{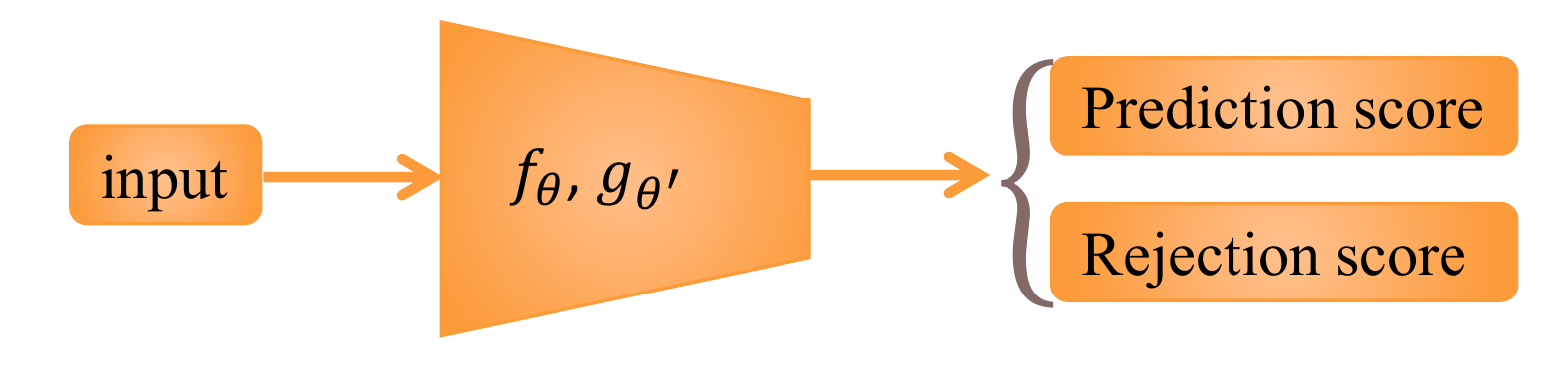}
\caption{A schematic overview of the prediction with reject option. Here, $f_\theta$ is the prediction function and $g_{\theta'}$ is the rejection or selection function}
\label{general_prediction_rejection}
\end{figure}

A typical supervised ML approach requires N training samples with corresponding labels such as $(x_1,y_1), (x_2,y_2)$,..., $(x_N,y_N)$ to learn a prediction function $f_\theta(x)$ which can predict labels for novel input. In the case of a neural network, $\theta$ represents the weights and biases of the network. During the training process, $f_\theta(x)$ is learned by minimizing an user-defined loss function $l(f_\theta(x),y)$ comparing the prediction score and the target values over training samples. Traditionally, this function will always produce a score for any vector with the same dimension as the training samples without considering whether the score is acceptable (reliable) or not. The general definition of selective prediction can be expressed as:
\begin{equation}
\label{prediction_with_reject_option}
(f_\theta,g_{\theta'})(x) \triangleq\left\{ \begin{array}{ll}
f_\theta(x), & \mbox{if } g_{\theta'}(x)=1\\
\mbox{refuse to predict}, & \mbox{if } g_{\theta'}(x)=0
\end{array}\right.
\end{equation}
where the selection or rejection function can be defined as:
\begin{equation}
\label{reject_function}
g_{\theta'}^\tau(x)=g_{\theta'}^\tau(x|\kappa_f)\triangleq\left\{ \begin{array}{ll}
0, & \mbox{if } \kappa_f(x)<\tau\\
1, & \mbox{if } \kappa_f(x)\geq\tau
\end{array}\right.
\end{equation}
here $\tau$ is a threshold that controls the quality of rejection, and $\kappa_f:x\rightarrow\mathbb{R}$ is a confidence rate function that calculates a metric (rejection score) reflecting the confidence in the prediction made $f_\theta$. A higher $\kappa_f$ value indicates higher acceptance of prediction. A schematic overview of the system is shown in Fig. \ref{general_prediction_rejection}.

\textit{Why do we need reject option?} Real-world data suffer from noise. For higher dimensional continuous data, it is difficult to estimate the exact boundary of any particular class and sometimes data from different classes overlap. Also, the model may get novel input which is far from training data distribution. These situations may lead to the incorrect output for even a well-trained model. Reject option is to make sure that the model stays away from predicting in case of any uncertain situation. Now, definitely the model could attain higher accuracy by rejecting most of the samples. Thus, the optimal choice is to reject as less as possible while maintaining high prediction accuracy.

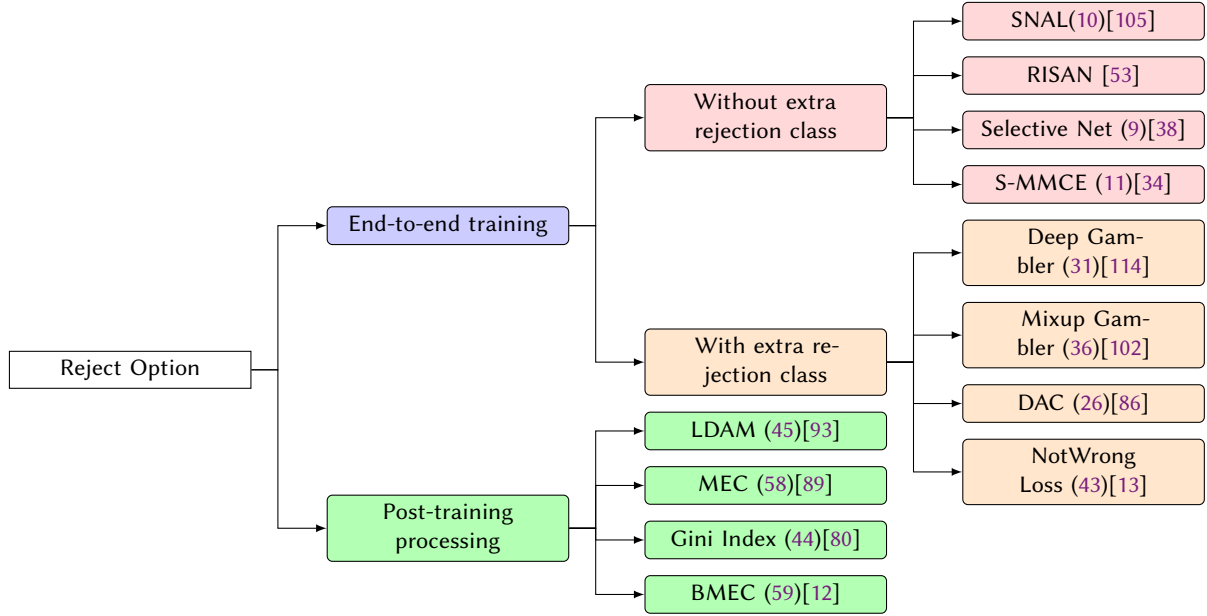
\begin{figure}
    \centering
    \tikzset{
    basic/.style  = {draw, text width=3cm, align=center, font=\sffamily, rectangle},
    root/.style   = {basic, rounded corners=2pt, thin, align=center, fill=green!30},
    onode/.style = {basic, thin, rounded corners=2pt, align=center, fill=green!60,,}, 
    pnode/.style = {basic, thin, align=center, rounded corners=2pt,fill=pink!60, 
    align=center},
    tnode/.style = {basic, thin, rounded corners=2pt,align=center, fill=orange!20, 
    align=center},
    xnode/.style = {basic, thin, rounded corners=2pt, align=center, fill=blue!20,},
    wnode/.style = {basic, thin, align=left, fill=pink!10!blue!80!red!10, text width=6.5em},
    edge from parent/.style={draw=black, edge from parent fork right}

}

\begin{forest} for tree={
    grow=east,
    growth parent anchor=west,
    parent anchor=east,
    child anchor=west,
    edge path={\noexpand\path[\forestoption{edge},->, >={latex}] 
         (!u.parent anchor) -- +(10pt,0pt) |-  (.child anchor) 
         \forestoption{edge label};}
}
[Reject Option, basic,  l sep=10mm,
    [Post-training processing, root,  l sep=10mm,
        [BMEC (\ref{BMEC})\cite{RN365},root][Gini Index (\ref{Gini_index})\cite{RN364},root][MEC (\ref{MEC})\cite{10.1109/ICASSP.2018.8461745},root][LDAM (\ref{LDAM})\cite{RN386},root]
 ]
    [End-to-end training, xnode,  l sep=10mm,
        [With extra rejection class, tnode, l sep=10mm,
        [NotWrong Loss (\ref{not_wrong_loss})\cite{RN14}, tnode][DAC (\ref{DAC})\cite{RN18}, tnode][Mixup Gambler (\ref{mixup_gambler})\cite{10.1007/978-3-030-86340-1_23}, tnode][Deep Gambler (\ref{deep_gambler})\cite{RN19}, tnode]
        ]
        [Without extra rejection class, pnode, l sep=10mm,
        [S-MMCE (\ref{SMMCE})\cite{10.48550/arxiv.2208.12084}, pnode][Selective Net (\ref{Selective_loss})\cite{RN337}, pnode][RISAN \cite{RN1}, pnode][SNAL(\ref{SNAL})\cite{9852645}, pnode]]
        ]
 ]
\end{forest}

    \caption{Classification of some common literatures with reject option}
    \label{fig:classification_reject_option}
\end{figure}

In this work, we try to comprehensively review the available work on the reject option for NNs. Broadly, we have encountered two types of strategies adopted by researchers:
\begin{enumerate}
    \item \textbf{Post-training processing}: Under this category, the rejection function $g_{\theta'}$ is learned after the training of the prediction function $f_{\theta}$ is already completed. A rejection or acceptance metric is formulated with the output of the trained $f_{\theta}$. Then a certain threshold of that metric is defined to decide whether a prediction should be accepted or rejected. Most papers explored in this study follow this strategy \cite{RN405,RN415,RN409,RN397,RN399,RN382,RN383,RN378,RN363,RN354,RN149,10.48550/arxiv.2204.13631,10.48550/arxiv.2207.10797}.
    \item \textbf{End-to-end training}: Many of the recent works follow this strategy where there is no post-training processing of the prediction output to generate the rejection metric. Under this category, there are two different approaches followed by the researchers: \begin{enumerate}
        \item  Here, instead of defining another metric, an extra class is appended to the available class category in the prediction function. Then, the whole model with extra class is developed using the training data. If the extra class holds the maximum score for any test sample, then this prediction of the input is said to be rejected \cite{RN19,RN18}.
        \item A novel loss function is formulated including both the prediction function and rejection function. NN is then trained by optimizing that novel loss which enables the simultaneous learning of the prediction function and the rejection function \cite{RN374,RN337}.
    \end{enumerate}
\end{enumerate}

\section{Loss Functions Directly incorporating rejection mechanism}
\label{sec:loss_func}
\begin{figure}
    \centering
    \begin{tikzpicture}[xscale=1.3]
  \def\t{0.056} 
    
  \draw[thick,evolcol!60!black,fill=hormcol!90,rounded corners=7,opacity=0.1]
    (0.0,-0.5) rectangle (3.5,0.5) ;
  \draw[thick,hormcol!60!black,fill=natalcol!90,rounded corners=7,opacity=0.2]
    (3.5,-0.5) rectangle (5,0.5);
  \draw[thick,ethcol!60!black,fill=ethcol!90,rounded corners=7,opacity=0.2]
   (5.0,-0.5) rectangle (8.0,0.5);
  \draw[thick,neurcol!60!black,fill=neurcol!90,rounded corners=7,opacity=0.2]
   (8.0,-0.5) rectangle (10,0.5);

 \node[neurcol!90!black,below=4,right=8,above left=2.5] at (10.7,-2.0) {2022};
 \node[neurcol!90!black,below=4,right=8,above left=2.5] at (8.1,-2.0) {2021};
 \node[neurcol!90!black,below=4,right=8,above left=2.5] at (5.95,-2.0) {2020};
  \node[neurcol!90!black,below=4,right=8,above left=2.5] at (3.5,-2.0) {2019};

  \draw[-to,branch,line width=14,opacity=0.3] (0,0) -- (10.5,0) node[right] {};

  \draw[branch,line width=2] (.2, 0.5) to[out=90,in=90,looseness=0.1]++ (0.0, 1.1)
    node[evolcol,above,align=center] {SelectiveNet\\Loss \cite{RN337}};
    \draw[branch,line width=2] (.3, -0.5) to[out=90,in=90,looseness=-0.1]++ (0.0, -1.1)
    node[evolcol,below,align=center] {DAC Loss \cite{RN18}};
      \draw[branch,line width=2] (1.7, 0.5) to[out=90,in=90,looseness=-0.1]++ (0.0, 0.4)
    node[evolcol,above,align=center] {Penalized Loss\\for SPS model \cite{RN154}};
        \draw[branch,line width=2] (1.8, -0.5) to[out=90,in=90,looseness=-0.1]++ (0.0, -0.4)
    node[evolcol,below,align=center] {Deep
Gambler\\Loss \cite{RN19}};
\draw[branch,line width=2] (3.2, 0.5) to[out=90,in=90,looseness=-0.1]++ (0.0, 1.1)
    node[evolcol,above,align=center] {Meta Loss \cite{RN374}};
    \draw[branch,line width=2] (3.3, -0.5) to[out=90,in=90,looseness=-0.1]++ (0.0, -1.1)
    node[evolcol,below,align=center] {Additive margin+\\Margined Unknown\\Loss \cite{RN358}};
\draw[branch,line width=2] (5.2, 0.5) to[out=90,in=90,looseness=-0.1]++ (0.0, 1.1)
    node[evolcol,above,align=center] {Cross-entropy\\with extra\\class \cite{Thulasidasan2021AnEB}};
    \draw[branch,line width=2] (5.8, -0.5) to[out=90,in=90,looseness=-0.1]++ (0.0, -0.5)
    node[evolcol,below,align=center] {NotWrong\\Loss \cite{RN14}};
    \draw[branch,line width=2] (6.5, 0.5) to[out=90,in=90,looseness=-0.1]++ (0.0, 0.5)
    node[evolcol,above,align=center] {RISAN\\Loss \cite{RN1}};
    \draw[branch,line width=2] (7.6, -0.5) to[out=90,in=90,looseness=-0.1]++ (0.0, -1.1)
    node[evolcol,below,align=center] {Mixup Gambler\\Loss \cite{10.1007/978-3-030-86340-1_23}};
    \draw[branch,line width=2] (7.8, 0.5) to[out=90,in=90,looseness=-0.1]++ (0.0, 1.1)
    node[evolcol,above,align=center] {PEBAL\\Loss \cite{RN3}};
    \draw[branch,line width=2] (8.8, 0.5) to[out=90,in=90,looseness=-0.1]++ (0.0, 0.5)
    node[evolcol,above,align=center] {SNAL \cite{9852645}};
    \draw[branch,line width=2] (9, -0.5) to[out=90,in=90,looseness=-0.1]++ (0.0, -0.51)
    node[evolcol,below,align=center] {S-MMCE \cite{10.48550/arxiv.2208.12084}};

\end{tikzpicture}

        \caption{Timeline for the novel loss functions associated with reject option.}
    \label{fig:loss_func_timeline}
\end{figure}
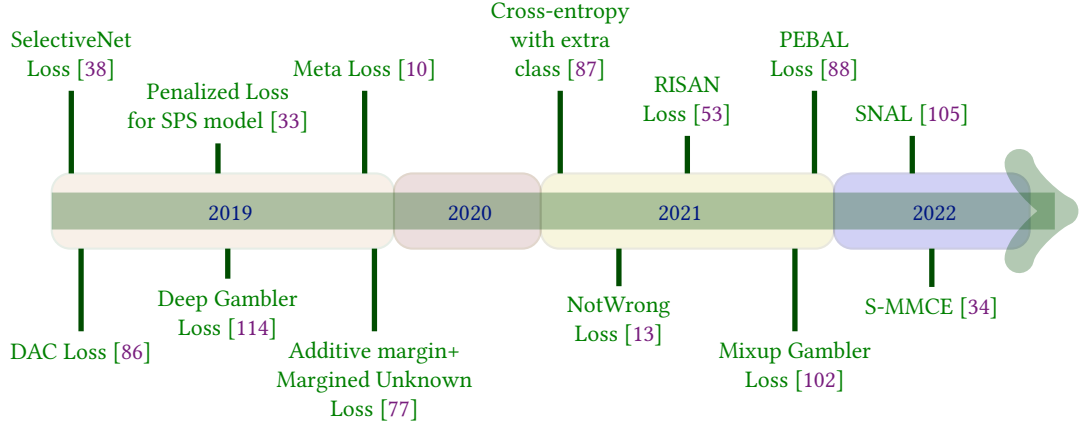

In this review, we found more than thirty cost functions corresponding to different training strategies and network formulizations. Here, we only wish to include those that are directly related to the rejection or selection of the network's output. A timeline for the novel loss functions is presented in Figure \ref{fig:loss_func_timeline}.

\subsection{Loss Function without Extra Rejection Class}
\subsubsection{Meta Loss Function}
Two NNs, one for the prediction model $f(x;\theta_f)$ and the other for the rejection model $g(x;\theta_g)$, were simultaneously learned in \cite{RN374}.  Here, $\theta_f$ and $\theta_g$ correspond to the weights of the prediction and rejection models respectively. The rejection model determines whether the output of the prediction model is accepted or discarded. The training method pushes the rejection model to produce a score greater than zero for the input that produces acceptable output in prediction model. To achieve this goal, a meta loss function $L(x,y)$ is designed to penalize both the wrong rejection by the rejection model $g(x)$ and the prediction error incurred by the prediction model $f(x)$. The convex meta loss function proposed by the authors is given below:
\begin{equation}
\label{meta_loss}
    L(x,y)=\max\left\{0,g(x;\theta_g)+l\left(f(x;\theta_f),y\right), \varsigma\left(1-g(x;\theta_g)\right)\right\}
\end{equation}

It can be seen from (\ref{meta_loss}) that for any input $x$, it produces positive value when the loss $l(f(x;\theta_f),y)$ is positive and the rejection model accepts the prediction $(g(x;\theta_g)>0)$. When the loss $l(f(x;\theta_f),y)$ is sufficiently low and the rejection model rejects the prediction $(g(x;\theta_g)\leq 0)$, a penalty consisting of rejection cost $\varsigma(1-g(x;\theta_g))$ is imposed. A high value of the hyper-parameter $\varsigma$ implies a high penalty for rejection. Hence, a very small number of samples will be rejected and vice versa. $L(x,y)$ is supposed to be minimized for all training samples during the training. Thus the minimization procedure can be expressed as:
\begin{equation}
\label{meta_loss_minimize}
    \theta_f^*,\theta_g^* = \underset{\theta_f,\theta_g}{argmin}\sum_{i=1}^NL(x_i,y_i)
\end{equation}

\subsubsection{SelectiveNet Loss}
\begin{figure}[!t]
\centering
\includegraphics[scale=0.45]{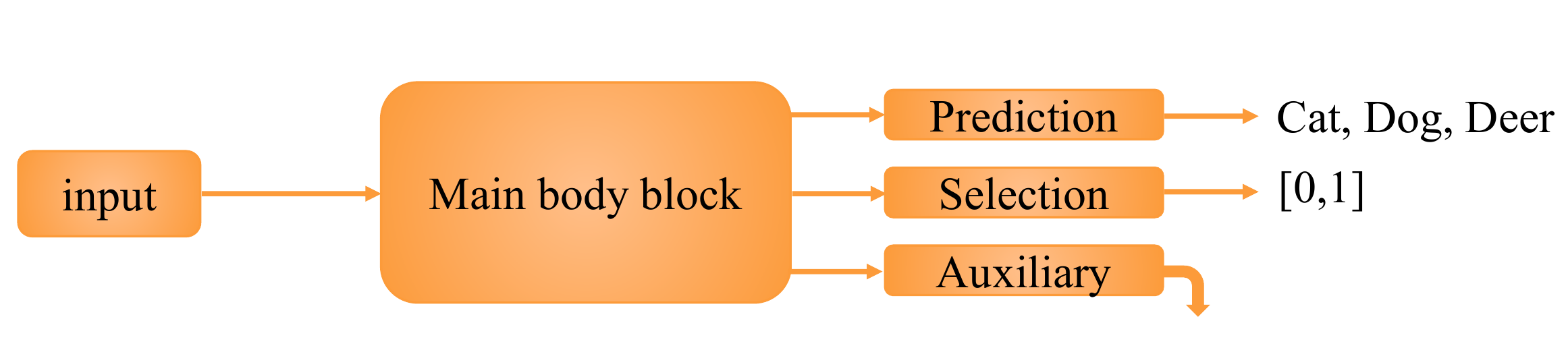}
\caption{SelectiveNet architecture (reproduced from \cite{RN337})}
\label{SelectiveNet}
\end{figure}

Yonatan Geifman and Ran El-Yaniv \cite{RN337} proposed an end-to-end optimization strategy for a neural network that provides a selective prediction (prediction with reject option). They termed their neural network architecture \textit{SelectiveNet}. It aims to optimize two functions; a prediction function $f(x)$ and a selection function $g(x)$ in a single DNN model. If $g(x)>0.5$, \textit{SelectiveNet} refrains from predicting the output for x. \textit{SelectiveNet} consists of three output heads for selection, prediction, and auxiliary prediction (see Fig. \ref{SelectiveNet}). Selection and prediction heads assist to implement the selection and the prediction functions correspondingly, while the auxiliary head enforces the network to learn the necessary features during the training period only. The schematic diagram of the SelectiveNet is given in Fig. \ref{SelectiveNet}.
The main body can be composed of any relevant architecture corresponding to the problem at hand such as recurrent NNs or long-short term memory networks, and convolution layers etc. The final layer of the selection head $g(x)$ consists of a single neuron from sigmoid activation and the last layer of the prediction head $f(x)$ can vary depending on the application such as softmax (classification) or linear (regression). The empirical selective risk $\hat{r}$ and the empirical coverage $\hat{\phi}$ are defined over a labeled set of $N$ samples $S_N$ as:
\begin{equation}
\label{risk}
    \hat{r}(f,g|S_N) \triangleq \frac{\frac{1}{N}\sum_{i=1}^N l(f(x_i),y_i).g(x_i)}{\hat{\phi}(g|S_N)}
\end{equation}
\begin{equation}
\label{coverage}
    \hat{\phi}(g|S_N)\triangleq \frac{1}{N}\sum_{i=1}^N g(x_i)
\end{equation}

To include the coverage constraint into the objective function, they used a variant of the Interior Point Method (IPM) \cite{POTRA2000281} which resulted in the following unconstrained selective loss function:
\begin{equation}
\label{IPM}
    L_{(f,g}) \triangleq \hat{r}(f,g|S_N)+\lambda\cdot\Psi(\mathbf{C}-\hat{\phi}(g|S_N))
\end{equation}
\begin{equation*}
    \Psi(a) \triangleq max(0,a)^2
\end{equation*}
where $\Psi$ is a quadratic penalty function, $\mathbf{C}$ is the expected coverage and $\lambda$ denotes the relative significance of the coverage constraint with respect to the risk value. 

The auxiliary head $h$ is also trained with the same loss as $f$:
\begin{equation}
\label{aux_loss}
    L_h = \hat{r}(g|S_N)=\frac{1}{N}\sum_{i=1}^N l(f(x_i),y_i)
\end{equation}
The final objective function formulated by the authors is the convex combination of selective loss $L_{(f,g)}$ from (\ref{IPM}) and the auxiliary loss $L_h$ from (\ref{aux_loss}) given as:
\begin{equation}
\label{Selective_loss}
    L_{SelectiveNet} = \alpha L_{(f,g)}+(1-\alpha).L_h
\end{equation}
The authors used 0.5 as the value of $\alpha$ for their experiment. They also suggested that without the auxiliary loss, the network only focuses on the $\mathbf{C}$ fraction of the training set before optimally learning the low-level feature which results in the over-fitting of a subset of the training set.

\subsubsection{Selective Net with Automatic Learning (SNAL)}
We notice from (\ref{IPM}) that coverage $\mathbf{C}$ is still needed to be specified by the expert. The optimal choice for $\mathbf{C}$ sometimes requires multiple trial-and-error using the whole dataset which may be very costly with large dataset. To dynamically learn it, Ye et al. proposed Selective Net with Automatic Learning (SNAL) in \cite{9852645} with the loss function shown in (\ref{SNAL}). Adding the extra $\lambda_C \cdot (1-\mathbf{C})$ term to (\ref{IPM}), loss in (\ref{SNAL}) ensures the optimal coverage for the dataset at hand.
\begin{equation}
\label{SNAL}
    L_{(f,g,\mathbf{C}}) \triangleq \hat{r}(f,g|S_N)+\lambda\cdot\Psi(\mathbf{C}-\hat{\phi}(g|S_N))+\lambda_C \cdot (1-\mathbf{C})
\end{equation}
They also proposed a soft version of the loss where the selection function $g(x)$ that has discrete output \{0,1\} is replaced by confidence rate function $\kappa_f$ producing a continuous confidence score [0,1] of any sample being accepted making the loss in (\ref{SNAL}) continuous and differentiable. After dividing the dataset into accepted and rejected group, seperate models are trained using them.

\subsubsection{Selective Maximum Mean Calibration Error (S-MMCE)}
To make prediction preferably on the well-calibrated uncertainties, Fisch et al. proposed Selective Maximum Mean Calibration Error (S-MMCE) \cite{10.48550/arxiv.2208.12084} as optimization objective which is based on the Maximum Mean Calibration Error \cite{pmlr-v80-kumar18a}. Let V be the conditional random variable where $V:= f(x)|g(x)=1$ $\in [0,1]$, The S-MMCE can then be expressed as:
\begin{equation}
\label{SMMCE}
    S-MMCE=\bigl\|\mathbb{E}[|\mathbb{E}[y|V]-V|^q ] \varphi(V)\bigl\|^{1/q}_\mathcal{H}
\end{equation}
where $q\geq 1$ and a universal kernel $\mathcal{K}$ induce a reproducing kernel Hilbert space $\mathcal{H}$ with feature map $\varphi:[0,1]\to \mathcal{H}$.
Due to computational complexity in (\ref{SMMCE}), upper bound to S-MMCE is proposed as:
\begin{equation}
\label{SMMCEU}
    S-MMCE_u=\bigl\|\mathbb{E}[|y-V|^q \varphi(V) ]\bigl\|^{1/q}_\mathcal{H}=\Biggl\|\frac{\mathbb{E}[|y-f(x)|^q g(x) \varphi(f(x))]}{\mathbb{E}[g(x)]}\Biggl\|^{1/q}_\mathcal{H}
\end{equation}

Now using the kernel trick, a finite sample estimate of $S-MMCE^2_u$ over a dataset $S_N$ can be computed as:
\begin{equation}
\label{SMMCEU_hat}
    \hat{S-MMCE}^2_u=\Biggl(\frac{\sum_{i,j\in S_N}(y_i-p_i)^q (y_j-p_j)^q g(x_i)g(x_j) \mathcal{K}(p_i,p_j)}{\sum_{i,j\in S_N}g(x_i)g(x_j)}\Biggl)
\end{equation}

To make the objective function differentiable, $g(x)$ is replaced by the confidence rate function $\kappa_f$ and the denominator in (\ref{SMMCEU_hat}) is removed simplifying the final loss function. To prevent $\kappa_f$ from collapsing to 0, a logarithmic regularization is utilized. Thus, the final loss function becomes:
\begin{equation}
\label{SMMCEU_loss}
    L_{S-MMCE}=\lambda_1 \Bigl(\sum_{i,j\in S_N}(y_i-p_i)^q (y_j-p_j)^q \kappa_f(x_i)\kappa_f(x_j) \mathcal{K}(p_i,p_j)\Bigl)-\lambda_2 \sum_{i\in S_N} \log \kappa_f(x_i)
\end{equation}
where hyper-parameters $\lambda_1,\lambda_2 \geq 0$.

\subsubsection{Additive margin Loss + Margined Unknown Loss}
 Shiraishi et al. in \cite{RN358} utilized \textit{additive margin 
softmax} method \cite{8331118} to formulate their loss function. In this method, the output of the penultimate layer is normalized $(||\mathbf{f}||=1)$ and used as a feature vector $\mathbf{f}$. If the bias term of the last layer is set to zero and the weights of the final layer for $j^{th}$ class $\mathbf{w_j}$ is normalized $(\mathbf{||w_j||}=1)$, the final logit can be calculated as the cosine similarity between $\mathbf{f}$ and $\mathbf{w_j}$ as :
\begin{equation}
\label{cos_similar}
    \Psi_{i,j} = \mathbf{w_j}.\mathbf{f_i} 
\end{equation}

Now, we can define the additive margin loss with margin $m$ as:
\begin{equation}
\label{loss_ams}
    L_{ams}=-\frac{1}{N}\sum_{i=1}^N log\frac{\exp^{a.(\Psi_{i,y_i}-m)}}{\exp^{a.(\Psi_{i,y_i}-m)}+\sum_{j=1,j\neq y_i}^c \exp^{a.\Psi_{i,j}}}
\end{equation}
where \textit{a} is the scale factor controling the magnitude of the exponential and $c$ is the available number of classes in dataset. In their training process, they also utilized a dataset outside of the target dataset which is generally termed "known unknown" samples. To include this extra dataset, they proposed \textit{margined unknown loss} which helps improve the performance of the model in rejecting unknown.
\begin{equation}
\label{loss_mul}
    L_{mul}=-\frac{1}{|D_U|}\sum_{i=1}^{|D_U|} log\frac{\exp^{a'.(1-m')}}{\exp^{a'.(1-m')}+\sum_{j=1}^c \exp^{a'.\Psi_{i,j}}}
\end{equation}
where $|D_U|$ is the set of known unknown training examples, $a'$ is the scale factor, and $m'$ is the margin hyper-parameter which enables the model to distinguish between novel class and registered class. The final loss is defined as:
\begin{equation}
\label{loss_mul_ams}
    L=L_{ams}+\alpha L_{mul}
\end{equation}
where $\alpha$ is the balancing weight that decides the relative significance of the two loss terms. In their experiment, the authors used 1.0 for $\alpha$.

\subsubsection{RISAN Loss}
Authors in \cite{RN1} proposed `Robust Instance Specific Deep Abstention Network' (RISAN) for a binary classifier. For the simultaneous learning of the prediction function $f\mathbf(x)$ and the rejection function $g\mathbf(x)$, they resorted to utilize double sigmoid loss function \cite{Shah2020OnlineAL} as given below:
\begin{equation}
\label{eq:double_sigmoid}
    L_{ds}(yf(\mathbf{x}),g(\mathbf{x}))= 2d\times\sigma(f(\mathbf{x})-g(\mathbf{x}))+2(1-d)\times\sigma(f(\mathbf{x})+g(\mathbf{x}))
\end{equation}
where $d$ represents the cost of rejection, $\sigma(a)=(1+\exp(\gamma a))$ denotes the sigmoid function with $(\gamma>0)$. For the detailed theoretical properties of this function, we refer the interested readers to the original work in \cite{RN1}. 

\begin{figure}[!t]
\centering
\includegraphics[scale=0.4]{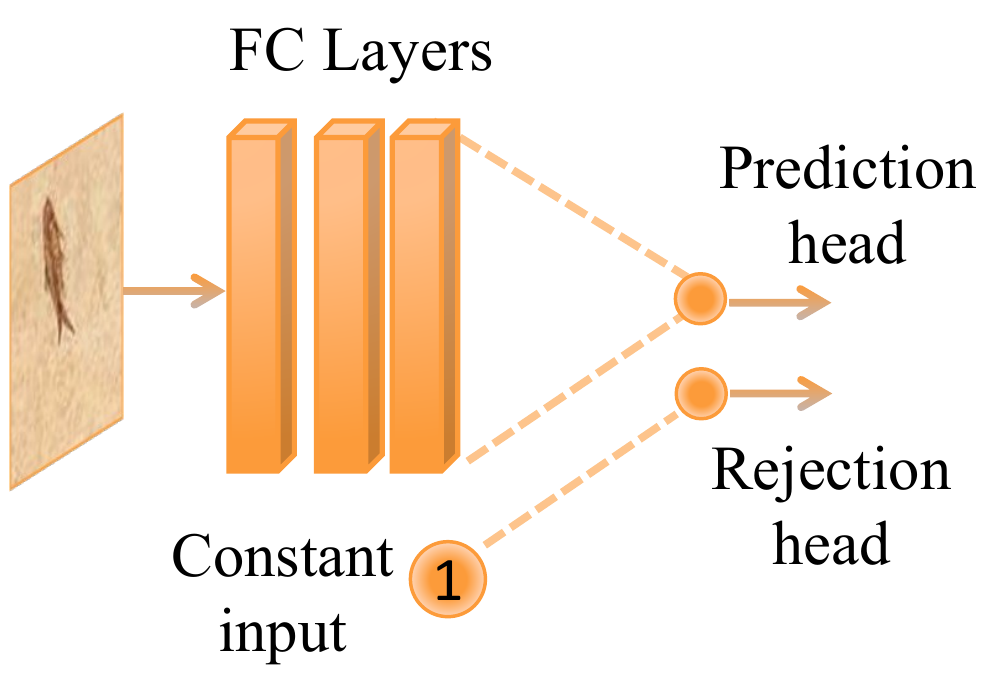}
\caption{RISAN architecture with input-independent rejection function (reproduced from \cite{RN1})}
\label{RISAN_input_independent}
\end{figure}

\begin{figure}[!t]
\centering
\subfloat{\includegraphics[scale=0.5]{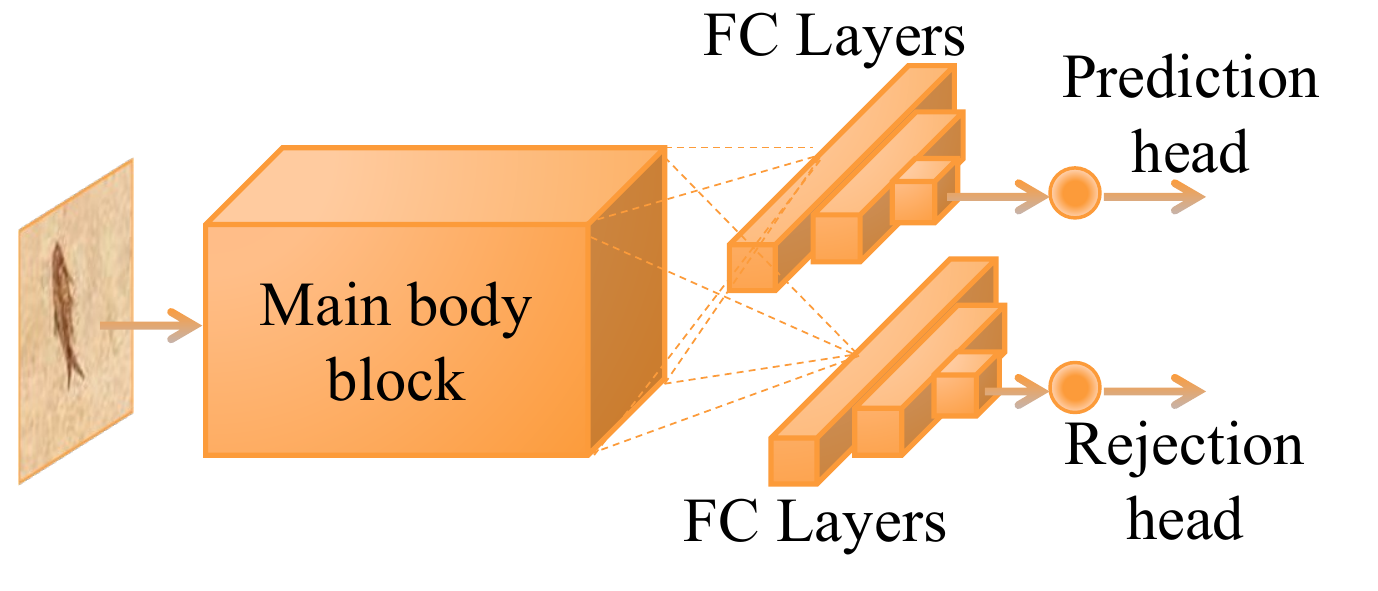} \label{RISAN_1}}
\hfil
\subfloat{\includegraphics[scale=0.5]{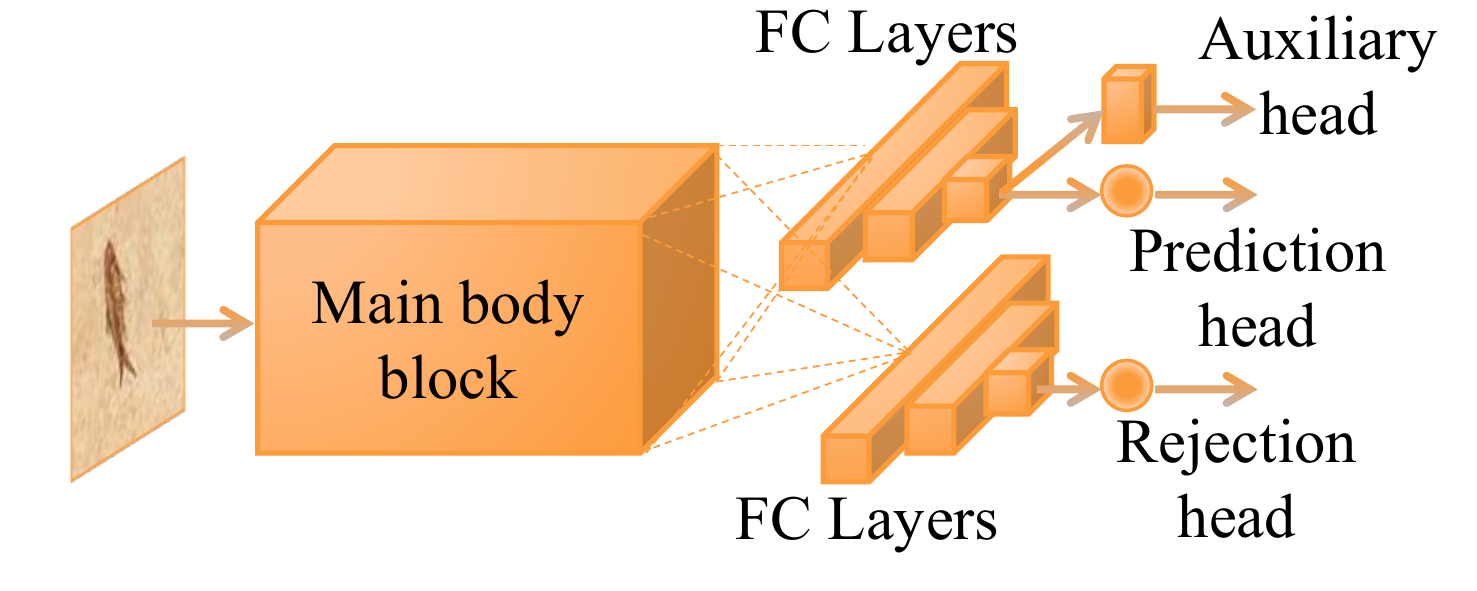} \label{RISAN_2}}
\caption{RISAN architecture with input-dependent rejection function (reproduced from \cite{RN1})}
\label{RISAN_input_dependent}
\end{figure}

Input independent RISAN architecture is shown in Fig. \ref{RISAN_input_independent} where the rejection function has the same value for all inputs. The prediction head tries to learn the appropriate prediction function $f\mathbf(x)$ while the rejection head learns the rejection function $g(\mathbf(x)$ parameter. Input-dependent RISAN architecture is shown in Fig. \ref{RISAN_input_dependent} where the rejection head is also fed the input from the main block through its own fully connected layers. Another architecture on the right side of Fig. \ref{RISAN_input_dependent} shows an additional auxiliary head that enforces the learning of relevant intermediate features in the initial steps. With the auxiliary head, the authors suggested learning the rejection and the prediction function optimizing a convex combination of the double sigmoid loss $(L_{ds})$ and the cross-entropy loss $(L_{ce})$ from the auxiliary head.
\begin{equation}
\label{L_c}
    L_c=\alpha L_{ds}+(1-\alpha) L_{ce}
\end{equation}

\subsubsection{NotWrong Loss for Regression Analysis}
Elizabeth and Barnes \cite{RN13} also proposed a variation of the previous loss function for regression analysis.
\begin{equation}
\label{not_wrong_loss_regression}
    L(\mathbf{x}_i)=-q_i\log p_i-\alpha\log q_i
\end{equation}
where $p_i$ denotes the value of the probability density function of a normal distribution having mean $\mu_i$ and standard deviation $\sigma_i$, $p_i=\mathcal{N}(y_i,\mu_i,\sigma_i)$ and $q_i$ is the weight of prediction, defined as:
\begin{equation}
\label{q_i_CAN}
    q_i=\min\Biggl(1.0,\biggl[\frac{\kappa}{\sigma_i}\biggl]^2\Biggl)
\end{equation}
where $\kappa$ is a data-dependent scale which can be defined as $\kappa=\mathcal{P}_m$ and $\mathcal{P}_m$ denotes the $m^{th}$ percentile of the predicted validation $\sigma$. The above loss increases $\sigma$ for samples with higher inference difficulty. Training of this CAN consists of two stages:
\begin{itemize}
    \item Spin-up: the CAN is trained without abstention for the first few epochs. At the end of the stage, $\mathcal{P}_m$ is estimated for different $m$ ranging from 10 to 90.
    \item Abstention training: Now, the already trained CAN continues to learn with the loss in \ref{not_wrong_loss_regression} and with $\kappa$ updated using $\mathcal{P}_m$ from the previous stage. During this stage, $\alpha$ is also updated accordingly.
\end{itemize}

\subsubsection{Penalized Loss for SPS model}
To include abstention, a penalized loss minimization framework is proposed by Feng et al. for training the selective prediction-set (SPS) model \cite{RN154}. The prediction function $f(x)$ and the decision (rejection) function $g(x)$ are learned simultaneously within the framework as shown below:
\begin{equation}
\label{SPS_abstention}
    \underset{f,g}{\min}\frac{1}{N}\sum_{i=1}^N\Bigl\{-\log f(y_i|x_i)g(x_i)+\delta\bigl(1-g(x_i)\bigl)\Bigl\}+\lambda_0 \frac{1}{N}\sum_{i=1}^N -\log f(y_i|x_i)+\lambda_1\int_{x\in\mathcal{D}} g(x) dx
\end{equation}
The first part of the objective function is termed as \textit{adaptively truncated loss}, the second part as \textit{information borrowing penalty}, and the third part as \textit{uniform acceptance penalty}. $\lambda_*$ are the scale factor for the respective penalty terms and $\delta$ denotes the penalty for abstention that is set based on prior knowledge. The first part incorporates the prediction loss and the abstention loss in a mutually exclusive manner. The \textit{information borrowing penalty} estimates the prediction loss over the whole training dataset encouraging the model to learn from samples that are previously abstained. $\lambda_0$ controls the amount of information learned from abstained examples. Finally, the \textit{uniform acceptance penalty} integrates the rejection function with respect to the Lebesgue measure over the data domain, enforcing the model to abstain from predicting less familiar samples. It incurs a lower penalty for the region with higher data density. Combining these three components, the final decision is influenced by the expected loss at any input $x$, the data density at $x$, and the penalty parameters $\lambda_0$ and $\lambda_1$.

\subsubsection{Metric Learning Loss}
Authors in \cite{RN366} proposed a new cost function, which they termed \textit{Metric Learning Loss}, to train their deep-RBF network as given below:
\begin{equation}
\label{J_ml}
    J_{ML}=\sum_{i=1}^N\Bigl(d_{y_i}(x_i)+\sum_{j\notin{y_i}}max\bigl(0,\lambda-d_j(x_i)\bigl)\Bigl)
\end{equation}
where $\lambda>0$ and $y_i$ corresponds to the true class of the input $x_i$. The first term in the loss function tries to decrease the distance between the $i^{th}$ input and its corresponding cluster centroid in the target feature space. With the help of hinge loss formulation, the second term pushes the distance of the sample from all the incorrect centroids outside of the margin $\lambda$ and at the same time prevents the function from collapsing. It is interesting to note that all the incorrect classes inside the margin produce a loss in function. During inference, decisions regarding rejection and prediction are directly taken based on the distances of a sample from cluster centroids.  Authors show that the classification accuracy of the network doesn't change significantly for a wide range of $\lambda$ making it easier to choose a proper value for it. 
To show that their loss function is equivalent to negative-log-likelihood, they also propose a soft version of their loss function as given below:
\begin{equation}
\label{J_softml}
    J_{SoftML}=\sum_{i=1}^N\biggl(d_{y_i}(x_i)+\sum_{j\notin{y_i}}log\Bigl(1+\exp\bigl(\lambda-d_j(x_i)\bigl)\Bigl)\biggl)
\end{equation}

\subsection{Loss Function with Extra Rejection Class}
In this section, the most well-known loss functions with extra rejection class are discussed.
\subsubsection{DAC Loss}
All loss functions discussed so far enforce NN models to reject or abstain on prediction during inference time. Thulasidasan et al. \cite{RN18} introduced a novel loss function that enables a model to reject samples (confusing or noisy samples) during training time, thereby increasing performance on non-abstained samples. This enables the model to be used as an effective data cleaner. The model, named Deep Abstaining Classifier (DAC), can learn features associated with the noisy samples even in the case of structured noise.  

For a given $x$, if $p^{(j)}=p_\theta(y=j|x)$ (the probability of $j^{th}$ class given x), then cross-entropy loss becomes $L_{cross-entropy}=-\sum_{j=1}^c y^{(j)} \log p^{(j)}$ for any sample where $y^{(j)}$ is the ground truth. To enable their model to achieve abstention during training, they introduced $c+1^{th}$ output $p^{(c+1)}$ to include the probability of abstention. They proposed the modified cross-entropy loss function for any sample $x_i$ as follows:
\begin{equation}
\label{DAC}
    L_{DAC}(x_i)=(1-p_i^{(c+1)}) \left( - \sum_{j=1}^c y_i^{(j)} \log \frac{p_i^{(j)}}{1-p_i^{(c+1)}}\right)+\alpha\log \frac{1}{1-p_i^{(c+1)}}
\end{equation}
where the first term is the adjusted cross-entropy, the second term enforce the penalty for abstention and  $\alpha$ is the scale factor for the penalty $(\alpha\geq0)$. In the absence of abstention $(p^{(c+1)}=0)$, (\ref{DAC}) becomes usual the cross-entropy loss. If $\alpha$ is very high, $p^{(c+1)}$ becomes very low, regenerating the standard cross-entropy loss. If $\alpha$ is very low, the lower penalty for abstention drives the model to abstain on everything with impunity, and the adjusted cross-entropy becomes very small. Between these two extreme situations, the model chooses to abstain on samples depending on how much cross-entropy error it produces while learning the ground truth compared to the abstention penalty.

To increase the abstention pre-activation during gradient descent, $\frac{\partial L_{DAC}}{\partial a_{c+1}}$ should be less than zero. To follow this constraint, the authors demonstrated that $\alpha$ must maintain the threshold: $\alpha < (1-p_i^{(c+1)}) \Bigl(- \log \frac{p_i^{(j)}}{1-p_i^{(c+1)}}\Bigl)$ where $p_i^{(j)}$ is the prediction corresponding to $j^{th}$ class for sample $x_i$. In their work, $\alpha$ is auto-tuned following this constraint. For the first L epochs, the model is trained without any abstention. At the beginning of the L+1 epoch, $\alpha$ is given a very small value enforcing abstention for most of the samples. As the training progresses, $\alpha$ is increased in every epoch up to a pre-defined final value. The whole process is shown in Algorithm \ref{alg:alpha_DAC}.
\begin{algorithm}
\caption{An algorithm to tune $\alpha$ (Reproduced from \cite{RN18})}\label{alg:alpha_DAC}
\begin{algorithmic}
    \Require Total iter. $(T)$, current iter. $(t)$, total epoch $(E)$, current epoch $(e)$, abstention free epoch $(L)$, $\alpha$ init scale factor $(\rho)$, final $\alpha$ $(\alpha_{final})$ and cross entropy over true classes $(L_{cross-entropy}(p_i))$, mini-batch size $(M)$
    \Ensure auto-tuning of $\alpha$ during training process
    \State Initialize $set_{\alpha}=False$
    \For{$t:=$0 to T}
    \If{$e<L$}
    \State $\beta = \frac{1}{M}\sum_{i=1}^M(1-p_i^{(c+1)})L_{cross-entropy}(p_i)$
    \If{$t=0$}
    \State $\tilde{\beta}=\beta$ (initialize moving average)
    \EndIf
    \State $\tilde{\beta}\leftarrow (1-\mu)\tilde{\beta}+\mu\beta$
    \EndIf
    \If{$e=L$ and \textbf{not} $set_{\alpha}$}
    \State $\alpha := \tilde{\beta}/\rho$
    \State $\delta_{\alpha}:=\frac{\alpha_{final}-\alpha}{E-L}$
    \State $set_{\alpha}=True$
    \State $Update_{epoch}=L$
    \EndIf
    \If{$e>Update_{epoch}$}
    \State $\alpha\leftarrow\alpha+\delta_{\alpha}$
    \State $Update_{epoch}=e$
    \EndIf
    \EndFor
\end{algorithmic}
\end{algorithm}

\subsubsection{Cross-Entropy with Extra Class}
Like the previous work, authors in \cite{Thulasidasan2021AnEB} introduced an extra class to their model but trained the model with the cross-entropy loss function rendering the optimization problem as:
\begin{equation}
\label{DAC_optim}
    \underset{\theta}{\min} \{\mathbb{E}_{(x,y) \thicksim \mathcal{D}_{in}}[-\log p_{\theta}(y=\hat{y}|x)]+\mathbb{E}_{(x,y) \thicksim \mathcal{D}_{out}}[-\log p_{\theta}(y=c+1|x)]\}
\end{equation}
where $\mathcal{D}_{out}$ denotes the set of all the outlier samples that are labeled as $(c+1)^{th}$ class. The advantage of the method over the other methods is the absence of any additional hyper-parameter. But the downside is the requirement of an outlier dataset which might not be well-defined and not always available.

\subsubsection{Deep Gambler Loss}
Ziyin et al. \cite{RN19} proposed a deep abstention classifier named ``Deep Gambler'' that is encouraged by the portfolio theory. The modern portfolio theory is an investment strategy in finance that helps investors assign their assets to businesses (stocks) while ensuring maximum profit return with minimum risk. Let's assume a stock market with $c$ stock is represented by a vector $\mathbf{x}=\{x_1, x_2, ..., x_c\}$ where  $x_i$ is defined as the ratio of the price of the $i^{th}$ stock at the beginning of the day to the price at the end of the day. The price vector can be considered as a vector of random variables generating from a joint distribution $\mathbf{x} \thicksim P(\mathbf{x})$. The investment (portfolio) in the different stocks of the stock market can be shown as $\mathbf{b}=(b_1, b_2, ..., b_c)$ where $b_i\geq0$ and $\sum_i b_i=1$. Therefore, the relative wealth at the end of the day can be expressed as $S=\mathbf{b}^T\mathbf{x}=\sum_i b_ix_i$. The speed at which the wealth increases, termed `doubling rate', can be expressed as:
\begin{equation}
\label{doubling_rate}
    W(\mathbf{b},P)=\int \log_2\Bigl(\mathbf{b}^T\mathbf{x}\Bigl) dP(\mathbf{x})
\end{equation}

Instead of considering the stock market, we will focus on horse races now. The only difference is that only one horse can win, and a horse either wins or loses. Thus, the outcome vector (previously price vector) for any race becomes one-hot vector $\mathbf{x(j)}= \{0, ...,0, 1, 0, .., 0\}$ where 1 is in the $j^{th}$ entry. Let's assume that we bet on m horses, and the probability of winning for $i^{th}$ horse is $p_i$, and the return on the betting is $o_i$ per dollar if the $i^{th}$ horse wins and 0 if loses. The portfolio $\mathbf{b}$ has the same constraint discussed before. If horse j wins, the relative wealth at the end of the race will be $S(\mathbf{x}(j))=b_j o_j$. The relative wealth after n races will be $S_n=\prod_{i=1}^n S(\mathbf{x}_i)$. We can treat races as a batch of samples considering the independence of the relative wealth on the order of the occurrence of the result of each race. Therefore, the doubling rate for any horse race becomes:
\begin{equation}
\label{doubling_rate_horse_race}
    W(\mathbf{b,p})=\mathbb{E}\log_2 (S)=\sum_{i=1}^c p_i \log_2(b_i o_i)
\end{equation}

If we take $o_i=1$ and $b_i$ as the post-softmax output of our model, then $W$ becomes standard cross-entropy loss. We want to invest a portion of our resources in the race while keeping the rest safe to minimize risk. That means, we can bet on $(c+1)$ category where $(c+1)^{th}$ is the abstention category with return $o_{c+1}=1$. The relative wealth after the race is now $S(\mathbf{x}(j))=b_j o_j+b_{c+1}$ and the doubling rate is given as:
\begin{equation}
\label{doubling_rate_horse_race_abstention}
    W(\mathbf{b,p})=\sum_{i=1}^c p_i \log_2(b_i o_i+b_{c+1})
\end{equation}
This is the Gambler loss. But, the authors deployed a slight variation of (\ref{doubling_rate_horse_race_abstention}) in their experiment of the c-class classification problem as shown below:
\begin{equation}
\label{deep_gambler}
    L_{gambler}=-\sum_{i=1}^c p_i \log(\hat{p}_i+\frac{1}{o}\hat{p}_{c+1})
\end{equation}
where $\hat{p}$ is the output of the NN, $\hat{p}_{c+1}$ denotes the rejection score, and $o$ is the rejection reward which should follow the constraint $1<o\leq c$.

\subsubsection{Mixup Gambler Loss}
\begin{figure}[!t]
\centering
\includegraphics[scale=0.42]{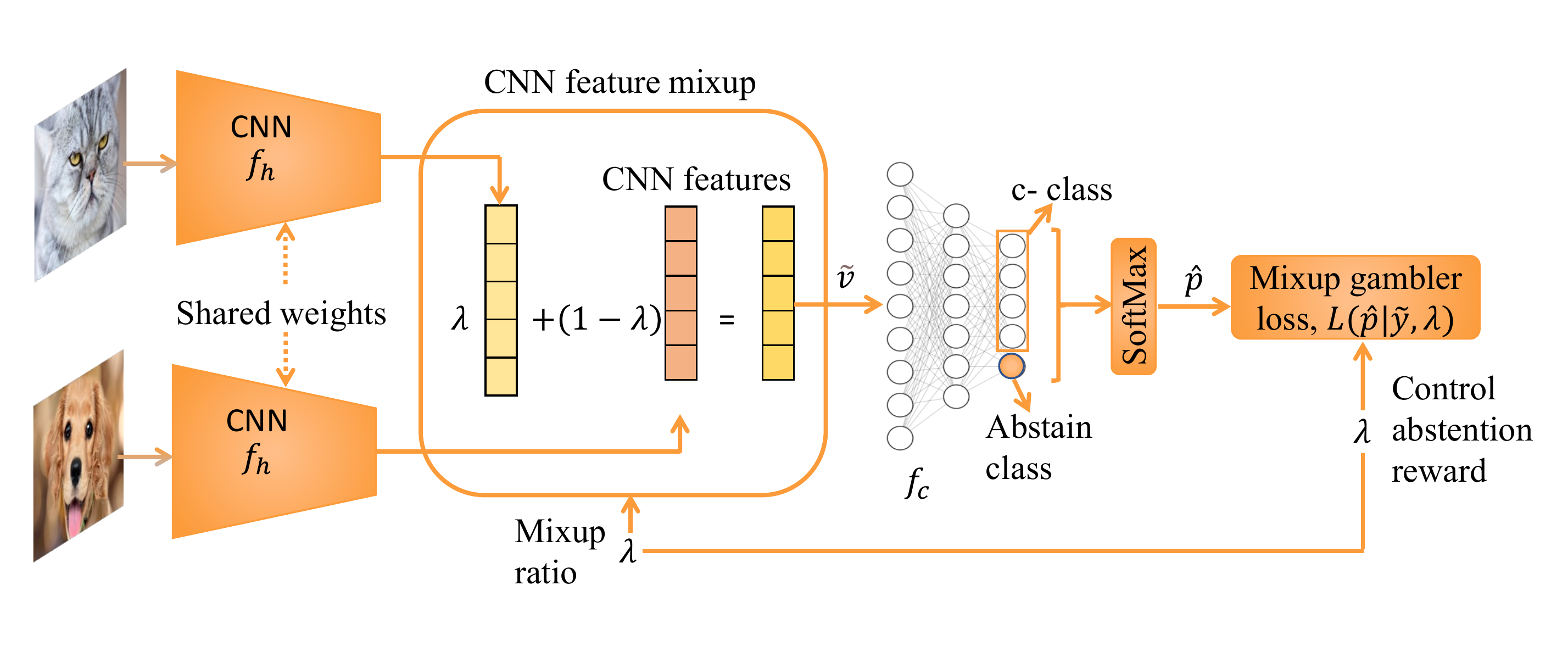}
\caption{Schematic view of mixup gambler (reproduced from \cite{10.1007/978-3-030-86340-1_23})}
\label{mixup_gambler_view}
\end{figure}

The rejection reward $(o)$ in the `deep gambler loss' discussed above has some limitations. Its value must be well-tuned for the optimum performance of the model using a validation dataset, making it dependent on the quality of the validation dataset. Input samples have different levels of inference difficulty. Thus, rejection rewards should also be different for different samples, but they are kept constant for all inputs in the `deep gambler loss'. Authors in \cite{10.1007/978-3-030-86340-1_23} addressed these two shortcomings and proposed a novel learning method where they tried to determine the rejection reward for each sample during training. They increase the reward for the hard input, to enforce the model to reject the prediction, while decreasing the rejection for easy input. To estimate $o$ effectively, they resorted to the mixup data augmentation strategy, where they assumed that the mixup ratio is directly related to the inference difficulty of samples. The schematic view of the proposed method is shown in Fig. \ref{mixup_gambler_view}.

Mixup data augmentation is defined as:
\begin{equation}
\label{mixup1}
    \mathbf{\tilde{x}}= \lambda \mathbf{x}^{(i)}+(1-\lambda)\mathbf{x}^{(j)}
\end{equation}
\begin{equation}
\label{mixup2}
    \mathbf{\tilde{y}}= \lambda \mathbf{y}^{(i)}+(1-\lambda)\mathbf{y}^{(j)}
\end{equation}
where $\mathbf{x}_i\in \mathbb{R}^D$ is the input features and $\mathbf{y}_i\in [0,1]^c$ is the target for $i^{th}$ sample. $\lambda$ is the mixup ratio where $\lambda\in [0,1]$. It is expected from the above equations that the difficulty of the augmented sample should be highest when $\lambda$ is 0.5 while it decreases when $\lambda$ deviates from 0.5. The rejection reward should also follow the same trend. Thus, the authors suggested using binary entropy function $\mathcal{H}(\lambda)$ as the rejection ratio. The authors also suggested that instead of pixel features (in the case of images), a mix of the intermediate features from CNN is necessary for the model to perform effectively. The prediction function $f$ can be considered to consist of two functions: the CNN feature extractor $f_h:\mathbf{x}\rightarrow\mathbf{v}$ and the classifier $f_c:\mathbf{v}\rightarrow\mathbf{y}$ where $\mathbf{v}$ is the hidden feature from the final CNN layer. Then the CNN feature mixup can be expressed as:
\begin{equation}
\label{mixup_CNN}
    \mathbf{\tilde{v}}= \lambda f_h(\mathbf{x}^{(i)})+(1-\lambda)f_h(\mathbf{x}^{(j)})
\end{equation}
\begin{equation}
\label{mixup_CNN_classifier}
    \mathbf{\hat{p}}= f_c(\mathbf{\tilde{v}})
\end{equation}
where $\mathbf{\hat{p}}$ follows the constraint $\sum_{i=1}^{(c+1)}\hat{p}_i=1$. Finally, the `mixup gambler loss' function can be shown as:
\begin{equation}
\label{mixup_gambler}
    L(\hat{\mathbf{p}}|\tilde{\mathbf{y}},\lambda)=-\sum_{i=1}^c \tilde{y}_i \log\biggl(\hat{p}_i+\frac{1}{1+\mathcal{H}(\lambda)(c-1)}\hat{p}_{c+1}\biggl)
\end{equation}
where $\tilde{\mathbf{y}}\in[0,1]^c$ is a mixup target vector that maintains $\sum_{i=1}^{c}\tilde{y}_i=1$ and $\mathcal{H}(\lambda)=-\lambda\log_2\lambda-(1-\lambda)\log_2(1-\lambda)$. The training strategy of the mixup gambler is summarized in Algorithm \ref{alg:mixup_gambler}. 
\begin{algorithm}
\caption{An algorithm to train mixup gambler model with a single mini-batch dataset (Reproduced from \cite{10.1007/978-3-030-86340-1_23})}\label{alg:mixup_gambler}
\begin{algorithmic}
    \Require mini-batch dataset $(\mathbf{X,Y})$, batch size m, CNN feature extractor $f_h$ having parameter $\theta_h$, fully connected classifier $f_c$ with parameter $\theta_c$, learning rate $\eta$, $\lambda_{min}(>0)$, $\lambda_{max}(<1)$, step size $\lambda_s$ for the increment of $\lambda$, mixup gambler loss function $L(\hat{\mathbf{p}}|\tilde{\mathbf{y}},\lambda)$
    \Ensure model parameters $\theta_h$, $\theta_c$
    \State $\mathbf{X',Y'}\leftarrow\mathbf{X,Y}$
    \State $\mathcal{L}\leftarrow 0$
    \For{$\lambda=\lambda_{min}$ to $\lambda_{max}$ \textbf{steps by} $\lambda_s$}
    \For{i=1 to $m$}
    \State $\mathbf{x,y}\leftarrow\mathbf{X[i],Y[i]}$
    \State $\mathbf{x',y'}\leftarrow\mathbf{X'[i],Y'[i]}$
    \State $\mathbf{\tilde{v}} \leftarrow \lambda f_h(\mathbf{x})+(1-\lambda)f_h(\mathbf{x'})$
    \State $\mathbf{\tilde{y}} \leftarrow \lambda \mathbf{y}+(1-\lambda)\mathbf{y'}$
    \State $\mathbf{\hat{p}} \leftarrow f_c(\mathbf{\tilde{v}})$
    \State $\mathcal{L} \leftarrow\ \mathcal{L}+L(\hat{\mathbf{p}}|\tilde{\mathbf{y}},\lambda)/m$
    \EndFor
    \EndFor
    \State $\theta_h\leftarrow\theta_h-\eta\nabla_{\theta_h}\mathcal{L}$
    \State $\theta_c\leftarrow\theta_c-\eta\nabla_{\theta_c}\mathcal{L}$
\end{algorithmic}
\end{algorithm}

\subsubsection{PEBAL Loss}
Tian et al. proposed pixel-wise energy-biased
abstention learning (PEBAL) \cite{RN3} for the segmentation task. The proposed cost function enables the simultaneous training of pixel-wise abstention learning (PAL) and energy-based model (EBM). To make the penalty of abstention in the PAL pixel variant, the EBM is suggested to be utilized to automatically estimate the penalty during the training for each pixel of each image. The schematic overview of the PEBAL method is shown in Fig. \ref{PEBAL_view}.

The PEBAL loss function is given below:
\begin{equation}
\label{PEBAL}
\begin{split}
l_{PEBAL}(\mathcal{D}^{in},\mathcal{D}^{out},\theta)=&\sum_{(\mathbf{x}^{in},\mathbf{y}^{in})\in\mathcal{D}^{in}} \Bigl(l_{pal}\bigl(\theta, \mathbf{y}^{in}, \mathbf{x}^{in},E_\theta(\mathbf{x}^{in})\bigl)+\lambda l_{ebm}^{in}\bigl(E_\theta(\mathbf{x}^{in})\bigl)+l_{reg}\bigl(E_\theta(\mathbf{x}^{in})\bigl)\Bigl)+\\ &\sum_{(\mathbf{x}^{out},\mathbf{y}^{out})\in\mathcal{D}^{out}} \Bigl(l_{pal}\bigl(\theta, \mathbf{y}^{out}, \mathbf{x}^{out},E_\theta(\mathbf{x}^{out})\bigl)+\lambda l_{ebm}^{out}\bigl(E_\theta(\mathbf{x}^{out})\bigl)+l_{reg}\bigl(E_\theta(\mathbf{x}^{out})\bigl)\Bigl)
\end{split}
\end{equation}
where $\lambda$ is a hyper-parameter, $\mathcal{D}^{in},\mathcal{D}^{out}$ denotes the inlier and outlier dataset, respectively, $\mathbf{y}^{in}\in\{0,1\}^{*\times c}$,  and $\mathbf{y}^{out}\in\{0,1\}^{*\times (c+1)}$ where class $(c+1)$ belongs to the pixel from anomalies in the training data, and $l_{pal}$ represents the PAL loss as defined below:
\begin{equation}
\label{PAL}
    l_{pal}(\theta,\mathbf{y},\mathbf{x},E_\theta(\mathbf{x}))=-\sum_{\omega\in\Omega}\log\biggl( f_\theta(y_\omega|x)_\omega+\frac{f_\theta(c+1|x)_\omega}{a_\omega} \biggl)
\end{equation}
where $a_\omega$ denotes the penalty for abstention for $\omega$ pixel. To define $a_\omega$, we need the inlier free energy $(E_\theta(\mathbf{x})_\omega)$ that is defined as:
\begin{equation}
\label{free_energy}
    E_\theta(\mathbf{x})_\omega=-\log\sum_{y\in\{1, 2, .., c\}} \exp\bigl( f_\theta(y|\mathbf{x})_\omega\bigl)
\end{equation}
From (\ref{free_energy}), we notice that low inlier free energy indicates the pixel being inlier and high inlier free energy indicates the pixel to be outlier. Now, the pixel-wise abstention penalty can be represented by $a_\omega=\bigl(-E_\theta(\mathbf{x})_\omega\bigl)^2$ where lower penalty means higher inlier energy, thus outlier pixel.
\begin{equation}
\label{l_ebm_in}
    l_{ebm}^{in}(E_\theta(\mathbf{x}))=\sum_{\omega\in\Omega}\bigl(\max(0,E_\theta(\mathbf{x})_\omega-m_{in})\bigl)^2
\end{equation}
The $l_{ebm}^{in}(.)$ term represents EBM loss which estimates the loss of the inlier samples with the free energy greater than the threshold $m_{in}$ while $l_{ebm}^{out}$ denotes the loss of the outlier samples with the free energy less than the threshold $m_{out}$. The margin loss in (\ref{l_ebm_in}) pushes the energy in the inlier samples to low values whereas (\ref{l_ebm_out}) pushes the energy of the outlier samples to high values thus creating an energy gap between them.

\begin{figure}[!t]
\centering
\includegraphics[scale=0.45]{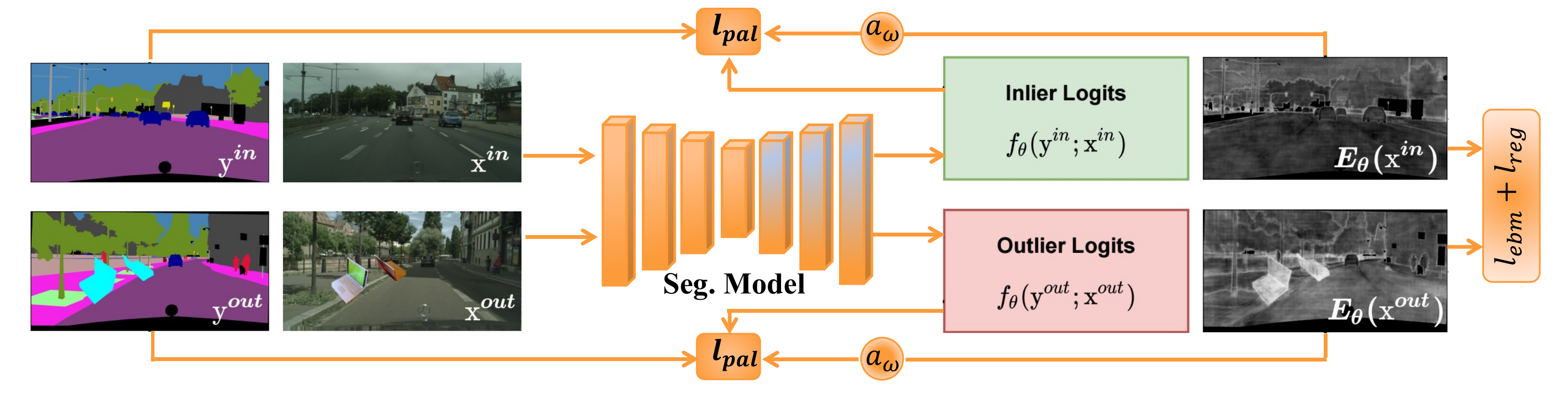}
\caption{Schematic overview of the PEBAL method (reproduced from \cite{RN3})}
\label{PEBAL_view}
\end{figure}

\begin{equation}
\label{l_ebm_out}
    l_{ebm}^{out}(E_\theta(\mathbf{x}))=\sum_{\omega\in\Omega}\bigl(\max(0,m_{out}-E_\theta(\mathbf{x})_\omega)\bigl)^2
\end{equation}
The $l_{reg}(.)$ term is the inlier free energy regularization loss that enforce the sparsity and the smoothness of the anomalous pixels and is defined as:
\begin{equation}
\label{l_reg}
    l_{reg}(E_\theta(\mathbf{x}))=\sum_{\omega\in\Omega}\beta_1|E_\theta(\mathbf{x})_\omega-E_\theta(\mathbf{x})_{\mathcal{N}(\omega)}|+\beta_2|E_\theta(\mathbf{x})_\omega|
\end{equation}
Here $\beta_1$ and $\beta_2$ are the weight hyper-parameters, and $\mathcal{N}(\omega)$ denotes the neighbouring pixels in horizontal and vertical directions.

\subsubsection{NotWrong Loss}
Elizabeth A. Barnes and Randal J. Barnes proposed `NotWrong loss' to train their controlled abstention network (CAN) \cite{RN14}. The formulation of the proposed loss enables the model to learn preferentially from more confident samples while refraining from difficult samples during training.
\begin{equation}
\label{not_wrong_loss}
    L_{NW}(\mathbf{x})=-\log(p^{(j)}+p^{(c+1)})-\alpha\log(\sum_{i=1}^c p^{(i)})
\end{equation}
where $p^{(j)}$ denotes the likelihood of the true class. The first term in the loss consists of the sum of the probability of the true class and that of the abstention class, forming the likelihood of "not wrong" prediction. The second term represents the penalty of abstention where $\alpha$ is the scale factor. Like DAC, `NotWrong loss' also enables the model to learn features related to the abstention class to some extent. The model can be made to learn a user-defined fraction of the training sample by adaptively controlling $\alpha$ during the training process.

\section{Post-Training Processing: Metric
\label{sec:post_process}
formulation and threshold determination}
To determine whether to accept or reject a prediction, we need metrics that can reflect the acceptance of the prediction and thresholds on them to decide the boundary between rejection and acceptance. The most general choice is to utilize the already available posterior probability from NNs \cite{RN414,RN370,RN401,RN403,RN394,RN393,RN349,RN386,RN346,RN384,RN341,RN342,RN362,RN148,article2019Othman,RN412}. The top prediction value and the relative difference between the top two prediction scores are used as the decision metric in \cite{RN414}, and thresholds are chosen considering the model performance on a validation dataset. Though only the top prediction is used in \cite{RN370,RN394}, they use different thresholds for different classes. For example, Gini Index (GI) is used for confidence estimation in prediction in \cite{RN364}.
\begin{equation}
\label{Gini_index}
    GI=1-\sum_{j=1}^c\left( p^{(j)}\right)^2
\end{equation}

Authors in \cite{RN386} mention a novel decision score named LDAM from \cite{He2011}, which has been observed to provide better results with an SVM classifier. Similarly, 
\begin{equation}
\label{LDAM}
    LDAM=\frac{\{\sum_{i=2}^c (\hat{p}^{(1)}- \hat{p}^{(i)})\}^2}{(c-1)^2 \Sigma_{12}}\\
\end{equation}
where $\hat{p}^{(i)}$ denotes the prediction score of any sample in descending order, $\mu_1=\hat{p}^{(1)}$, $\mu_2=\frac{1}{c- 1} \sum_{i=2}^c\hat{p}^{(i)}$, $\Sigma_1=0$,
$\Sigma_2=\frac{1}{c-1}\sum_{i=2}^c (\hat{p}^{(i)}- \mu_2)^2$, and $\Sigma_{12}=\frac{1}{2}\Sigma_2$.

Chagas et al. \cite{10.1117/12.2606273} used Helmholtz free energy as a scoring function for the prediction of a DNN. Helmholtz free energy can be defined as:
$E(x;f)=-T*\log\left(\sum_{j=1}^c \exp^{p^{(j)}/T}\right)$ where $T$ is the temperature scaling value.

The following formula demonstrates a novel uncertainty measurement proposed in \cite{RN360}.
\begin{equation}
\label{Uncertainty_metric}
    J(p)=\max(p)- \mu(p')- \sigma(p')
\end{equation}
where $p'=p- \max(p)$, $\mu(p')$, and $\sigma(p')$ are the mean and standard deviation of $p'$.

Zhang et al. \cite{RN356} utilized predictive entropy by implementing the MC-dropout method to estimate an uncertainty score for the rejection function.
\begin{equation}
\label{predictive_uncertainty}
    \mathcal{H}=-\frac{\sum_{i=1}^M\sum_{j=1}^c p_i^{(j)}\ln(p_i^{(j)})}{M}
\end{equation}
where $M$ is the MC-dropout samples per input. Entropy from the output of the softmax layer is used as an uncertainty measurement in \cite{Wang2018IDKCF}.

Authors in \cite{RN148} defined `credibility' and `confidence' metrics based on prediction values such as: $\mbox{credibility}=\max{p^{(j)}}; j\in {1,2,..,c}$ and 
\begin{equation}
\label{confidence_credibility}
    \mbox{confidence}=1-\mbox{second highest } p.
\end{equation}
They then used a linear combination of credibility and confidence as an evaluation function, such as $\kappa=a_1.\mbox{credibility}+a_2.\mbox{confidence}$. The optimal values for a and b are computed by minimizing the AURC (Area Under the Risk Coverage curve) with a grid search over [-1,1]. The threshold is chosen such that the empirical risk and coverage are almost equal to the accepted risk and coverage. 

The thresholds are chosen by solving an optimization problem where the goal is to achieve higher accuracy on non-rejected prediction while rejecting less than an acceptable number of the prediction \cite{RN370}. ROC curve was used in \cite{RN394} for optimal threshold estimation. In \cite{RN412}, a generalized effectiveness function $\zeta(d_c \eta_c,d_r \eta_r,d_e \eta_e)$ is proposed to be maximized to get the optimal threshold value under the constraints: $\frac{\partial\zeta}{\partial\eta_c}>0$, $\frac{\partial\zeta}{\partial\eta_r}<0$, $\frac{\partial\zeta}{\partial\eta_e}<0$ and $|\frac{\partial\zeta}{\partial\eta_r}|<|\frac{\partial\zeta}{\partial\eta_e}|$ where $d_c$, $d_r$ and $d_e$ are the cost of correct classification, rejection and wrong prediction respectively. $\eta_c$, $\eta_r$ and $\eta_e$ are the percentage of correct classification, rejection and misclassification, respectively. For detailed analysis, we refer the interested reader to \cite{RN412}.

The p-norm of the gradient of the penultimate layer is used as a novelty score in \cite{Rn334}. The gradients of the layers can be described as a familiarity index of an input with the model. Familiar input doesn't require the model to update itself and produces lower gradient values. During the test time, the model is not updated, but the gradients are calculated for the penultimate layer. Moreover, a new uncertainty measurement, called "Blend-var", is proposed in \cite{RN150}, where the variance of multiple outputs for a single input is determined. The MC-dropout model can predict multiple outputs for a single input utilizing different model architectures due to the dropout layers. In case of "Blend-var", different predictions come from augmenting the test sample (rotating, shifting etc for image). 

Authors in \cite{10.48550/arxiv.2207.07506} proposed a combined confidence score of $S_1$ and $S_2$ where $S_1$ can differentiate between correct and incorrect prediction within in-distribution and $S_2$ can distinguish between correctly predicted in-distribution and out-of-distribution samples. Finally, the combined score can be estimated as:
\begin{equation}
\label{SIRC}
    C(S_1,S_2)=-(S_1^{max}-S_1)(1+\exp (-b[S_2 - a])),
\end{equation}
where a,b are hyper-parameters, $S_1$ can be softmax score or entropy from (\ref{predictive_uncertainty}), and $S_2$ can be $l_1$-norm of the feature vector $\|z\|_1$ from \cite{NEURIPS2021_063e26c6} or the negative of the residual score $-\|z^{p^\perp}\|_2$ from \cite{10.48550/arxiv.2203.10807}. This combination is termed as Softmax Information Retaining Combination (SIRC).

A new class-sensitive reliability score based on cosine similarity is proposed in \cite{RN336}.
\begin{equation}
\label{cosine_reliability}
    s_l(p_i)=\frac{1}{N_c}\sum_{k=1}^{N_c}\cos(p_i,p_{c_k})-\frac{1}{N_f}\sum_{k=1}^{N_f}\cos(p_i,p_{f_k})
\end{equation}
where the output of a DNN is divided into correct and incorrect groups per class. For class label $l$, $N_c$ and $N_f$ are the total number of prediction samples in correct and incorrect (false) group, respectively. $p_{c_k}$ and $p_{f_k}$ are the prediction samples in correct and incorrect (false) group, respectively. First, correct and incorrect prediction sets are formed for each label using a portion of the dataset. Now, the reliability score for any input can be calculated using (\ref{cosine_reliability}). To estimate the best threshold, we need another portion of the dataset that also generates two prediction groups for each label. Now the thresholds for $l$ class can be expressed as: $t_{cl}=\max(s_l(p_c))$ and $t_{fl}=\max(s_l(p_f))$ where $p_c\in\mbox{correct prediction set}$, $p_f\in\mbox{incorrect prediction set}$.
Finally, the decision is made as follows:
\begin{equation}
\label{cosine_reliability_accept}
    s_l(p_i)\geq t_{cl},\mbox{ prediction accepted}
\end{equation}
\begin{equation}
\label{cosine_reliability_reject}
    s_l(p_i)\leq t_{fl},\mbox{ prediction rejected (outlier)}
\end{equation}
\begin{equation}
\label{cosine_reliability_abstein}
   t_{fl}< s_l(p_i)< t_{cl},\mbox{ prediction abstained}
\end{equation}

Authors in \cite{RN386,RN364} determined the best threshold by minimizing empirical risk $\hat{R}=w_r R+E$, where $R$ and $E$ represent the fraction of rejected and misclassified samples, and $w_r$ is the cost of rejection.

Geifman and El-Yaniv \cite{RN156} presented a binary search algorithm named "selection with guaranteed risk" (SGR), which enables the selection function to learn the optimal threshold while guaranteeing a prescribed risk with sufficient confidence. The process is summarized in Algorithm \ref{alg:SGR}. Let $B^* (\hat{r}_i,\delta ,S_N)$ be the solution of the following equation:
\begin{equation}
\label{risk_bound}
    \sum_{j=1}^{N.\hat{r}(f|S_N)}\left( \begin{array}{c} N \\ j \end{array} \right) b^j (1-b)^{N-j}=\delta
\end{equation}
Here $0<\delta<1$ and $\hat{r}(f|S_N)$ is the emprirical error for $f_\theta$. If $R(f|S_N)$ is the true error, then the risk is bounded by $\mathbf{Pr}_{S_N}\left\{R(f|S_N)>B^* (\hat{r}_i,\delta ,S_N)\right\}<\delta$. For detailed analysis, we refer the interested readers to \cite{RN156}.

\begin{algorithm}[!t]
\caption{SGR algorithm (Reproduced from \cite{RN156})}\label{alg:SGR}
\begin{algorithmic}
    \Require prediction function f, confidence-rate function $\kappa_f$, a small confidence parameter $\delta (\delta>0)$, target risk $r^*$, and training dataset $S_N$
    \Ensure threshold $\tau$, risk bound $b^*$
    \State Sort $S_N$ according to $\kappa_f(xi)$, $x_i \in S_N$
    \State $a_{min}=1$, $a_{max}=N$
    \For{i=1 to $\log_2(N)$} 
    \State $a=(a_{min}+a_{max})/2$
    \State $\tau=\kappa_f(x_a)$
    \State $\hat{r}_i=\hat{r}(f,g^\tau|S_N)$
    \State $b_i=B^* (\hat{r}_i,\delta / \lceil\log_2 N \rceil ,S_N)$; using \ref{risk_bound}
    \If{$b_i<r^*$}
    \State $a_{max}=a$
    \Else
    \State $a_{min}=a$
    \EndIf
    \EndFor
    \State $b^*=b_i$
    \State Return $b^*$ and $\tau$
\end{algorithmic}
\end{algorithm}

A confidence estimation of the probability values generated by a trained CNN is introduced in \cite{RN365}. The confidence score is defined by a zero-order smooth-step function using the posterior probability:
\begin{equation}
\label{conf_score}
    ConfScore(\mathbf{p^*},\mathbf{\beta})=\left\{ \begin{array}{rc}
    0, & \mathbf{p^*\beta^T}\leq \gamma_1\\
    \frac{\mathbf{p^*\beta^T}-\gamma_1}{\gamma_2-\gamma_1}, & \gamma_1<\mathbf{p^*\beta^T}< \gamma_2\\
    1, & \mathbf{p^*\beta^T}\geq \gamma_2
\end{array}\right.
\end{equation}
Here, $\gamma_1$ and $\gamma_2$ are the user defined hyper-parameters deciding the left and right edges of the scaled probability value, $\mathbf{p}^*=[p_1^*,p_2^*,...,p_c^*]$ is the posterior probability vector sorted in the descending order and $\mathbf{\beta}=[\beta_1, \beta_2,...,\beta_c]$ is a coefficient vector. This confidence estimation function can be considered a feed-forward network cascaded on top of the softmax layer of the CNN. In their study, they treated $\mathbf{\beta}$ as a learnable parameter that can be adjusted in a supervised way given the training data $D_{out}=\{(\mathbf{p}_1^*,\hat{y_1},\tilde{y_1}), (\mathbf{p}^*,\hat{y_2},\tilde{y_2}), ..., (\mathbf{p}^*,\hat{y_N},\tilde{y_N})\}$. Here, $\tilde{y_i}$ is the ground truth label and $\hat{y_i}$ is the estimated class from the softmax layer of the CNN using the following equation:
\begin{equation}
\label{softmax_output}
    y_i = \underset{y^{(j)}}{arg max} \{p(y^{(j)}|x_i)\}_{j=1, 2, ..., c}
\end{equation}
Then, they relabeled the dataset as $D_r=\{(\mathbf{p}_1^*,l_1), (\mathbf{p}_2^*,l_2), ..., (\mathbf{p}_N^*,l_N)\}$ where $l_i$ is calculated using the following equation:
\begin{equation}
\label{label_score}
    l_i=\left\{ \begin{array}{rc}
    1, & \tilde{y_i}=\hat{y_i}\\
    -1, & \tilde{y_i}\neq\hat{y_i}
\end{array}\right.
\end{equation}

To estimate the performance of the confidence score concerning the relabelled dataset $D_r$, a performance metric should be formulated. Wan et al. \cite{10.1109/ICASSP.2018.8461745} formulated the mean effective confidence (MEC) as:
\begin{equation}
\label{MEC}
    MEC=\frac{1}{N}\sum_{i=1}^N ConfScore(\mathbf{p^*},\mathbf{\beta}) * l_i
\end{equation}
From (\ref{label_score}) and (\ref{MEC}), it is obvious that a higher value of MEC is only possible when the correct classification has a higher confidence score and the wrong classification has a lower confidence score. It is also important to notice that MEC is sensitive to data imbalances in $D_r$ such as too many correct classifications or too many wrong classifications. To alleviate this problem, authors in \cite{RN365} defined balanced MEC (BMEC) as:
\begin{equation}
\label{BMEC}
    BMEC=\frac{1}{N_1}\sum_{i\in D_{r1}} ConfScore(\mathbf{p_i^*},\mathbf{\beta}) * l_i+\frac{1}{N_2}\sum_{j\in D_{r2}} ConfScore(\mathbf{p_j^*},\mathbf{\beta}) * l_j
\end{equation}
where $D_{r1}$ corresponds to correctly classified $(l=1)$ dataset only and $D_{r2}$ consists of incorrect classifications $(l=-1)$ only. $N_1$ and $N_2$ are the number of samples in $D_{r1}$ and $D_{r2}$, respectively. The optimal value of $\beta$ can be found by maximizing BMEC. The authors used the Genetic Algorithm (GA) for this case. The whole process is summarized in Algorithm \ref{alg:_beta_conf_score}.
\begin{algorithm}[!t]
\caption{An algorithm to obtain optimal $\hat{\beta}$ for ConfScore (Reproduced from \cite{RN365})}\label{alg:_beta_conf_score}
\begin{algorithmic}
    \Require The outcome of the CNN $D_{out}$, hyper-parameters $\gamma_1$ and $\gamma_2$
    \Ensure optimal value of $\beta$
    \State Construct $D_r$ from $D_{out}$ using equation \ref{label_score}
    \State Initialize $\beta$
    \While{GA is not converged} 
    \State Evaluate ConfScore for each sample in $D_r$ using equation \ref{conf_score}
    \State Calculate BMEC using equation \ref{BMEC}
    \State Update $\beta$ using BMEC as an objective function in GA
    \EndWhile
    \State Return $\hat{\beta}$
\end{algorithmic}
\end{algorithm}

To measure the best threshold for ConfScore from (\ref{conf_score}), the authors proposed a novel evaluation metric that is defined as:
\begin{equation}
\label{fit_score}
    \mbox{Fit}=\mbox{TAR}+\mbox{TRR}-1
\end{equation}
Before defining TAR and TRR, we argue that the decision of rejection function can be divided into true
acceptance (TA) cases, true rejection (TR) cases, false rejection (FR) cases, and false acceptance (FA) cases. Now, TRR and TAR can be expressed as:
\begin{equation}
\label{TAR}
    \mbox{TAR}=\frac{\sum\mbox{TA}}{\sum\mbox{TA}+\sum\mbox{FR}}
\end{equation}
\begin{equation}
\label{TRR}
    \mbox{TAR}=\frac{\sum\mbox{TR}}{\sum\mbox{TR}+\sum\mbox{FA}}
\end{equation}

In \cite{RN393}, Benso et al. deployed the covariance matrix adaptation evolution strategy (CMA-ES) to automate the process of evaluating optimum threshold parameters for rejection procedures of a neural network as this strategy does not require any preliminary hypothesis on the solution. Sensitivity and specificity were utilized in the formulation of the objective functions. The possibility of an uncertain outcome was considered in evaluating those metrics as shown in the equations.
\begin{equation}
    sens'=\frac{TP}{TP+FN+TP_{uncertain}}
    \label{sen}
\end{equation}
\begin{equation}
    spec'=\frac{TN}{TN+FP+FP_{uncertain}}
    \label{spec}
\end{equation}
where TP = True Positive, TN = True Negative, FP = False Positive, and FN = False Negative.

The three objective functions that the authors experimented with in their work are given below:
\begin{equation}
\label{SS}
    SS=(1-sens')+(1-sepc')
\end{equation}
\begin{equation}
\label{SS1}
    SS1=1-\frac{sens'.spec'}{sens'+spec'} 
\end{equation}
\begin{equation}
\label{SS2}
    SS2=1-(sens'.spec') 
\end{equation}
It is clear from the above equations (\ref{SS}-\ref{SS2}) that higher values of sens' and spec' decrease the value of the objective function. As CMA-ES is designed to minimize the value of the objective function, (\ref{SS}) indicates that sensitivity and specificity are treated separately, while (\ref{SS1}) and (\ref{SS2}) try to maximize both jointly. According to the authors, SS1 is the best function that incorporates the internal relationship between sensitivity and specificity.\par
Distance from the center of the clusters generated using high-level features from the DNN is used for decision making in \cite{RN17}. Here, one single class could consist of multiple clusters. For each class, the threshold is chosen to be the highest distance from the cluster centers corresponding to the class (only the lowest distance is considered as multiple distances are possible considering multiple clusters for each class). Distance metric $d_j(x_i)$ from (\ref{J_softml}) is used to make decisions in \cite{RN366} while the threshold is defined as :
\begin{equation}
\label{RBF_threshold}
    T=-\log\Bigl(\frac{-1+\sqrt{1+\exp(\lambda_{eval})}}{2\exp(\lambda_{eval})}\Bigl)
\end{equation}
where $\lambda_{eval}$ is another hyper-parameter different from the $\lambda$ in (\ref{J_softml}).

Mahalanobis distance is then used in \cite{DBLP:journals/corr/abs-2002-04205} to estimate the closeness of any sample from multi-variate Gaussian distributions that are generated from the pre-softmax activations termed as kernel activation vector (kav) from a trained DNN using the training data. To enable the operation scalable to a larger dataset and wider softmax layer, the authors assumed that the elements in the activation vector are linearly independent, thus the Mahalanobis distance becomes: $d_{ij}=\sqrt\frac{(\mbox{kav}_i-\mu_j)^2}{\sigma_j^2}$
Here, $d_{ij}$ denotes the distance of $i^{th}$ sample from the distribution of $j^{th}$ class with mean $\mu_j$ and standard deviation $\sigma_j$.

To determine the threshold on the distance, authors mention that the squared Mahalanobis distances
follow $\chi^2_{df}$ distribution with a degree of freedom \textit{df} being equal to the dimension of the distribution. They also noticed that at a very high degree of freedom, $\chi^2_{df}$ becomes an approximation of a normal distribution with mean=\textit{df} and standard deviation=2.\textit{df}. Now, let us assume that $\mathcal{N}_{kav}(\mu_1,\sigma_1^2)$ and $\mathcal{N}_{kav}(\mu_2,\sigma_2^2)$ be the kav distributions of the top-2 classes, the $1^{st}$-quantile decision rule ($\leq\mu+\sigma$) is used to determine whether a sample is outlier or not. If $d_{i,joint}^2\leq\mbox{df}+2.\mbox{df}\rightarrow\mbox{outlier}$, otherwise, in-distribution sample. Here, $d_{i,joint}$ is the Mahalanobis distance if $i^{th}$ test sample from the joint kav distribution of top-2 classes defined as:
\begin{equation}
\label{joint_mahalanobis}
d_{i,joint}=\sqrt\frac{(\mbox{kav}_i-(\mu_1^2+\mu_2^2))^2}{\sigma_1^2+\sigma_2^2}
\end{equation}

In the case of the in-distribution sample, it is checked whether the distances from each of the top two class distributions are within a certain percentage (learned empirically) of each other. If it is, then the sample is declared to be uncertain.

\begin{figure}[!t]
\centering
\includegraphics[scale=0.5]{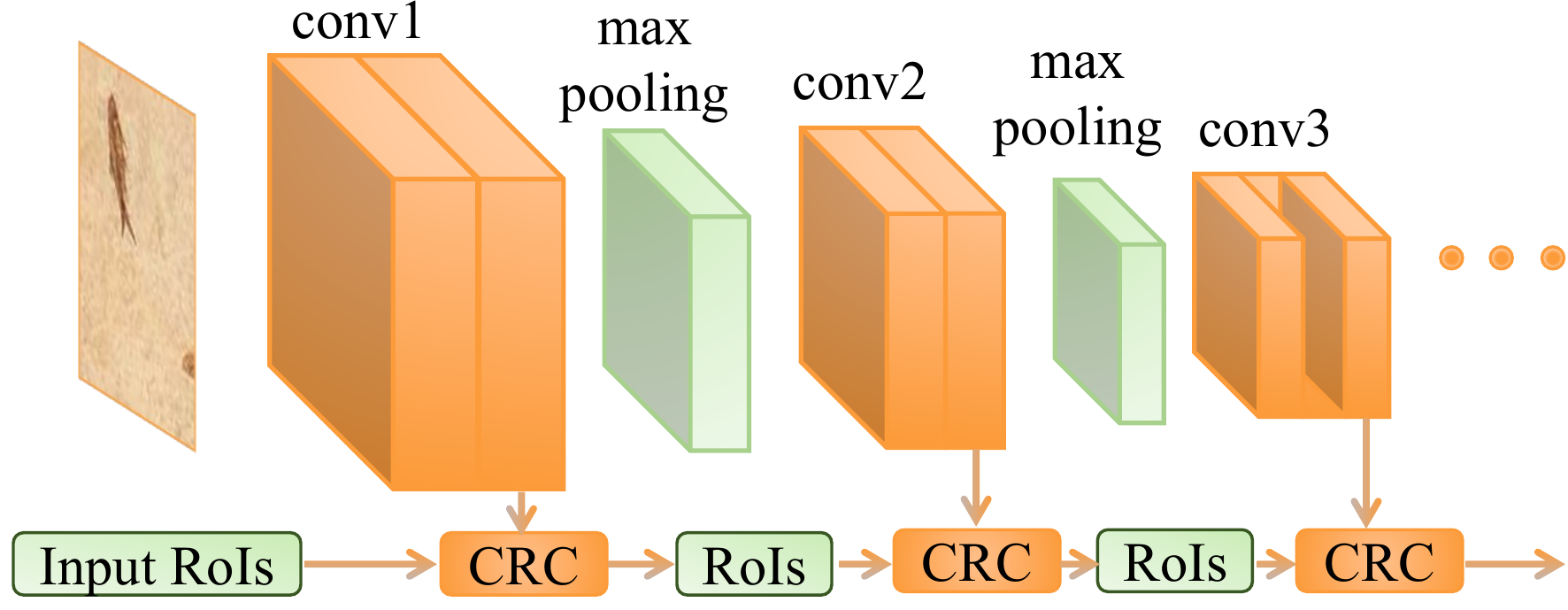}
\caption{Schematic overview of CRC. Here, RoI means the region of interest pooling of features. (reproduced from \cite{RN381})}
\label{CRC}
\end{figure}

Yang et al. \cite{RN381} proposed to utilize the features from all the convolution layers of a CNN to build rejection classifiers in a cascaded manner termed cascaded rejection classifiers (CRC), where non-rejected inputs are passed to the next classifier in the cascade. The general structure is shown in Fig. \ref{CRC}.

\begin{figure}[!t]
\centering
\includegraphics[scale=0.4]{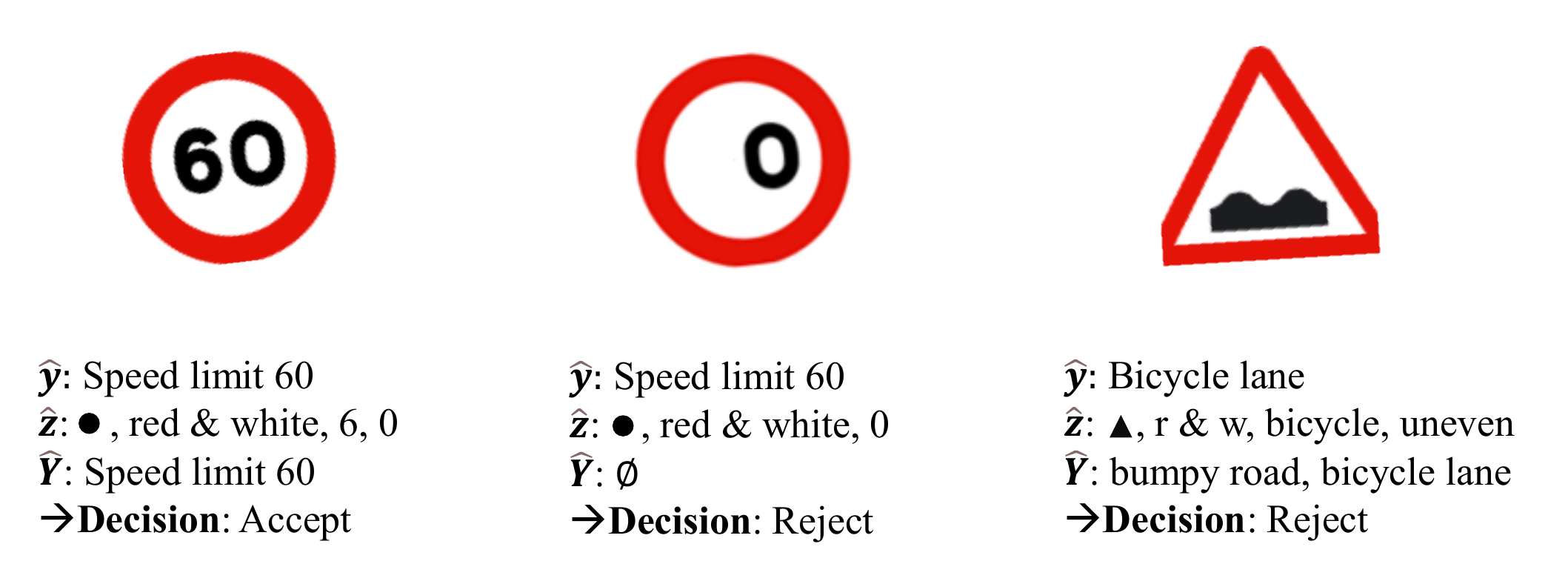}
\caption{Example of decision making using explainable attributes (reproduced from \cite{RN335})}
\label{ex_attributes}
\end{figure}

If we have features that we can understand and use to explain reasonably the correlation between the input and the outcome of a DNN, then the decision-making becomes very easy as conducted in \cite{RN339,RN344}. In a mobile robot localization problem, authors considered the possible route of the robot arriving at a destination from a particular starting point to estimate the decision made by a CNN. If the destination is not possible considering the route the robot has taken, the prediction of the CNN is rejected. Authors in \cite{RN335} estimated low-level explainable attributes using CNN layer features for reliable decision-making as shown in Fig. \ref{ex_attributes}.

\begin{figure}[!t]
\centering
\includegraphics[scale=0.5]{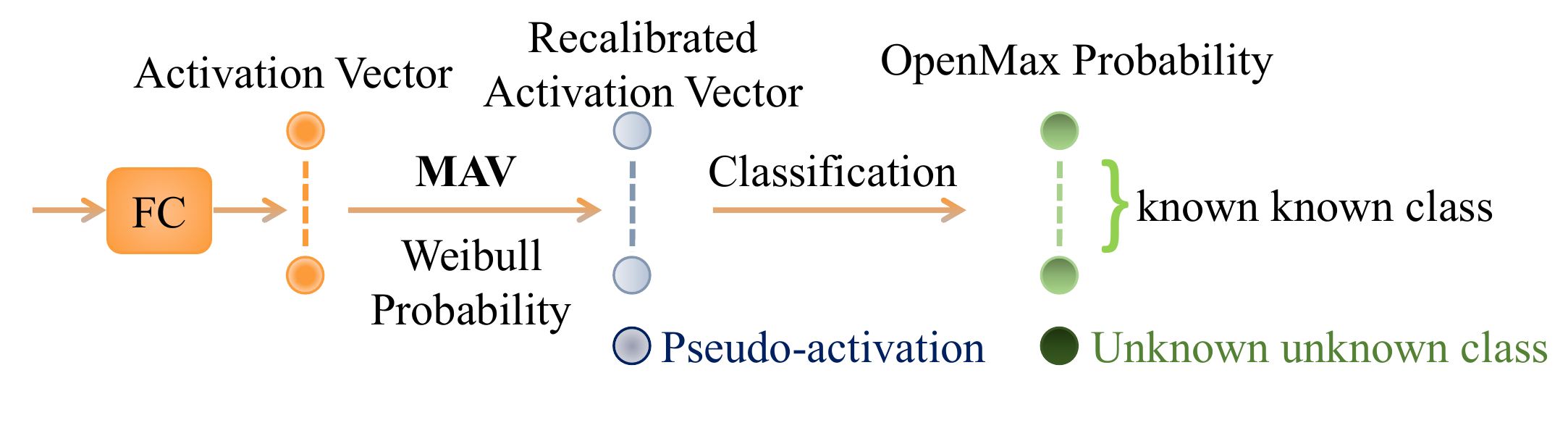}
\caption{Schematic view of the OpenMax layer (reproduced from \cite{RN351})}
\label{openmax}
\end{figure}

Liu et al. \cite{RN351} replaced the softmax layer of trained DNN with OpenMAX \cite{7780542} layer. The schematic diagram of OpenMax layer is shown in Fig. \ref{openmax}.

It has been observed that the output of the penultimate layer (termed as activation vector) of a DNN differs significantly among different classes: known or unknown. Thus, we can determine a mean Activation Vector (MAV) for each class over the correctly classified training samples only. The
Euclidean Distances $D(,)$ between their AV and their corresponding class’s MAV, are calculated to fit the Weibull distribution of their own class. Finally the
Weibull Cumulative Distribution Function (CDFWeibull) can be estimated based on the following equation:
\begin{equation}
\label{weibull_dist}
    CDF_{Weibull}(D(f(x),MAV_i);\rho_i)=\exp\left(- \left(\frac{(D(f(x),MAV_i))}{\eta_i}\right)^{a_i}\right); i=1,2,..., c
\end{equation}
where $\rho_i=(a_i,\eta_i)$ is calculated from the correctly classified training data of $i^{th}$ class and $x$ is the input corresponding to $i^{th}$ class. At testing time, the following equations (\ref{argsort})-(\ref{y_hat_new}) are used to compute a pseudo-activation vector for extra (rejection) class.
\begin{equation}
\label{argsort}
    s(j)=argsort(y^{(j)}); j=1,2,..., c
\end{equation}

\begin{equation}
\label{w_s}
    w_{s(j)}(x)=1-\frac{1-\alpha}{\alpha}CDF_{Weibull}(D(f(x),MAV_i);\rho_i); i=1,2,..., \alpha
\end{equation}
\begin{equation}
\label{y_hat}
    \hat{\mathbf{y}}=\mathbf{y}\circ\mathbf{w(x)}
\end{equation}

\begin{equation}
\label{y_hat_new}
    \hat{y}^{(c+1)}=\sum_{j=1}^c y^{(j)}(1-w_j(x))
\end{equation}
where $1\leq\alpha\leq c$, $\circ$ implies element-wise multiplication, and $s(j)$ denotes the index of the elements in the activation vector (AV) in descending order. Now, the new posterior probability can be calculated as:
\begin{equation}
\label{posterior_softmax_extra_class}
  p^{(j)}=P(C_j|x)=\frac{\exp(\hat{y}^{(j)})}{\sum_{i=1}^{c+1}\exp(\hat{y}^{(i)})}; j=1,2,..., c+1
\end{equation}

\begin{figure}[!t]
\centering
\includegraphics[scale=0.52]{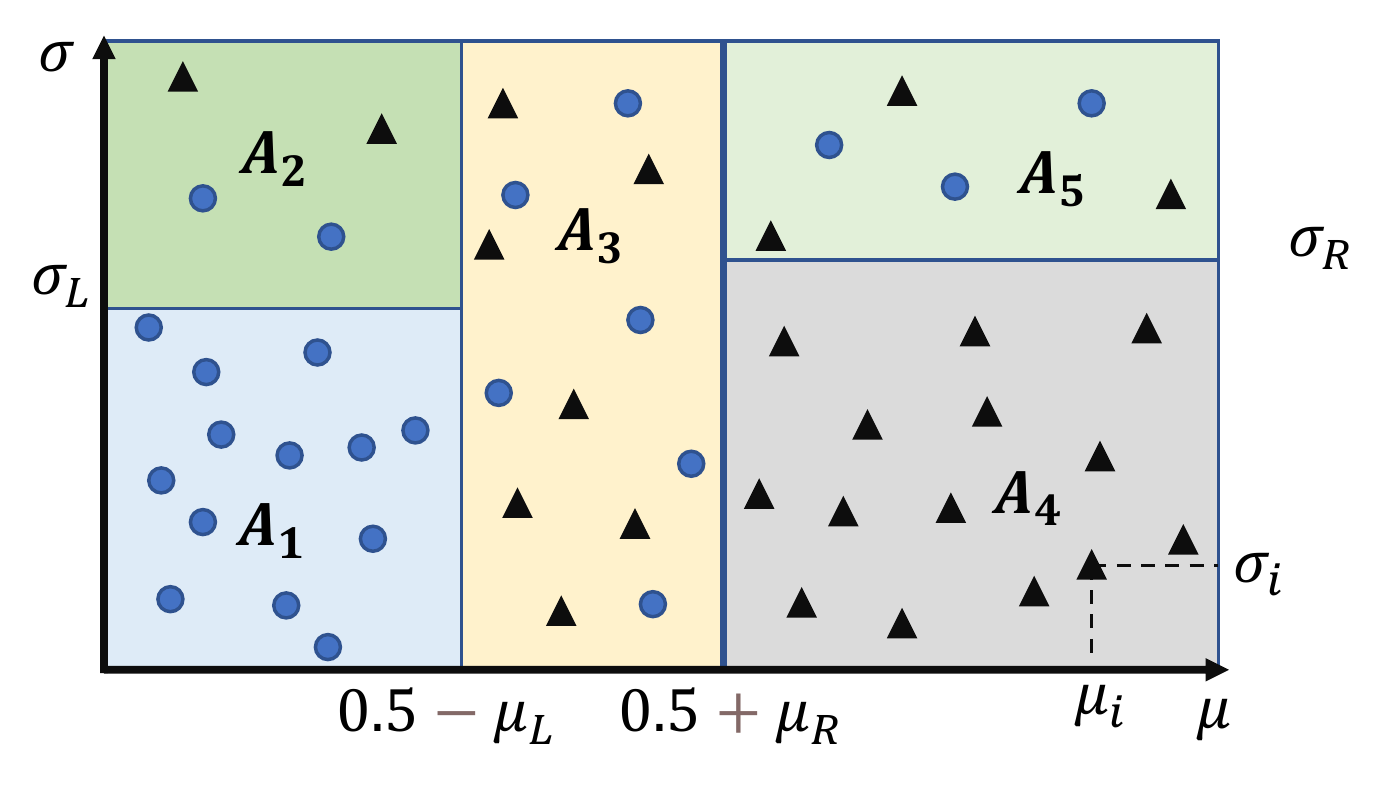}
\caption{Mean-variance space in the MIPSC model (reproduced from \cite{RN146})}
\label{MIPSC}
\end{figure}

Authors in \cite{RN350} proposed to utilize nonconformity score for rejection function. A nonconformity function can be shown as:
\begin{equation}
\label{nonconformity_func}
  h(z_i,\zeta)=\triangle[f(x_i),y_i]
\end{equation}
where $z_i=(x_i,y_i)$, $\zeta=z_1, z_2,...$, and $\triangle$ is a special function to measure prediction error. The margin error function is used in this work as shown below:
\begin{equation}
\label{margin_error}
  \triangle[f(x_i),y_i]=\underset{y\neq y_i}{\max}(\hat{p}_f (y|x_i)-\hat{p}_f (y_i|x_i))
\end{equation}
where $\hat{p}_f (y|x_i)$ denotes the probability score obtained from prediction function $f$ for input $x_i$ corresponding to class $y$. Now, nonconformity scores can be calculated using the following steps:
\begin{enumerate}
    \item Divide the dataset into four subsets: $D_{train},D_{val},D_{cal},D_{test}$ where $|D_{cal}|=q$
    \item Use $D_{train},D_{val}$ to train and validate $f$
    \item Estimate nonconformity scores $\{\alpha_1, \alpha_2, ..., \alpha_q\}= \{ h(z_i,\zeta):\zeta=D_{train}, z_i\in D_{cal}\}$
\end{enumerate}
For a sample $x_i\in D_{test}$, the decision is made by producing a prediction region $\Gamma_i^\epsilon \subseteq Y$ (Y is a class label set) as follows:
\begin{enumerate}
    \item Set a significance level $\epsilon\in(0,1)$
    
    \item For every class $\tilde{y}\in Y$: \begin{enumerate}
        \item Tentatively label $x_i$ as $(x_i,\tilde{y}_i)$
        \item Calculate $\alpha_i^{\tilde{y}}=h[(x_i,\tilde{y}_i),D_{train}]$
        \item Estimate $p_i^{\tilde{y}}=\frac{|\{z_j\in D_{cal}:\alpha_j\geq\alpha_i^{\tilde{y}}\}|+1}{q+1}$
        \item Determine $\Gamma_i^\epsilon=\{\tilde{y}\in Y:p_i^{\tilde{y}}>\epsilon\}$
    \end{enumerate}
\end{enumerate}
The prediction is rejected if p-values corresponding to all classes are less than $\epsilon$ in the region $\Gamma_i^\epsilon$. The largest p-value greater than $\epsilon$ indicates the label of the sample. For the binary classification task, authors in \cite{RN146} applied MC-dropout \cite{pmlr-v48-gal16} to estimate the predictive mean and variance for each sample and plotted them in 2-D space for characterizing the posterior distribution of the whole data space. To determine the optimal decision and rejection area in this 2-D space, they utilized the "Mixed Integer Programming" optimization method. The space in the mean-variance space is divided into five decision areas as shown in Fig. \ref{MIPSC}.
$A_1$ and $A_2$ denote the regions corresponding to positive and negative classification, respectively. $A_4$ and $A_5$ represent rejection region with high predictive variance (high model uncertainty). Boundaries for these regions might not be associated with each other due to the class specific pattern in the data. The region close to 0.5 predictive mean is another rejection region denoted as $A_3$. Our objective is to find boundaries $(\mu_L,\mu_R,\sigma_L,\sigma_R)$ of these regions by solving the MIP formulation. Let's develop the constraints for samples in different regions.
\begin{table}
\caption{Notation for MIPSC}
\begin{tabular}{r|l}
Symbol & Meaning\\
\hline
$y_i$ & True label \\
$p_i$ & Indicator of positive class decision \\
$n_i$ & Indicator of negative class decision\\
$r_i$ & Indicator of rejection decision\\
$\mu_i$ & Predictive mean \\
$\sigma_i$ & Uncertainty measure \\
$\mu_L$ & Left mean boundary for rejection \\
$\mu_R$ & Right mean boundary for rejection \\
$\sigma_L$ & Upper uncertainty border for positive class \\
$\sigma_R$ & Upper uncertainty border for negative class \\
$\mathcal{A}_{L_i}$ & Left area indicator\\
$\mathcal{A}_{R_i}$ & Right area indicator\\
$\mathcal{A}_{DL_i}$ & Down left area indicator\\
$\mathcal{A}_{DR_i}$ & Down right area indicator\\
$r_{Cap}$ & Rejection capacity\\
$\xi$ & Accuracy without rejection\\
$\epsilon$ & Very small constant\\
$\mathcal{M}$ & Very large constant
\end{tabular}

\label{tab:definition}
\end{table}

Constraint (\ref{MIP_1}) holds for the samples that reside on the right side of $A_3$ region such that $i\in A_4\cup A_5$ while constraint (\ref{MIP_2}) and (\ref{MIP_3}) characterize the samples in region $A_4$, $i\in A_4$. Similarly, Samples on the left side of the $A_3$ region follow the constraint (\ref{MIP_4}) such that $i\in A_1\cup A_2$ while constraint his region and (\ref{MIP_6}) hold for the samples in region $A_1$, $i\in A_1$. Notations are defined in Table \ref{tab:definition}.

\begin{equation}
\label{MIP_1}
    \mu_i>0.5+\mu_R\iff\mathcal{A}_{R_i}=1
\end{equation}
\begin{equation}
\label{MIP_2}
    \sigma_i<\sigma_R \iff \mathcal{A}_{DR_i}=1
\end{equation}
\begin{equation}
\label{MIP_3}
    \mathcal{A}_{R_i}+\mathcal{A}_{DR_i}>1 \iff n_i=1
\end{equation}
\begin{equation}
\label{MIP_4}
    \mu_i<0.5-\mu_L \iff \mathcal{A}_{L_i}=1
\end{equation}
\begin{equation}
\label{MIP_5}
    \sigma_i<\sigma_L \iff\mathcal{A}_{DL_i}=1
\end{equation}
\begin{equation}
\label{MIP_6}
    \mathcal{A}_{L_i}+\mathcal{A}_{DL_i}>1 \iff p_i=1
\end{equation}
All samples must produce one of the three output: positive class, negative class and rejection such that $p_i+n_i+r_i=1$. Finally, we also want to control the number of rejections following the constraint:
\begin{equation}
\label{MIP_7}
    (\sum_{i=1}^{N}r_i)\leq r_{Cap}
\end{equation}
Therefore, the main goal is to reduce the wrong prediction as well as the number of rejection. Following the goal and the above constraint, the MIP model becomes:
\begin{equation}
\label{MIP_8}
    \underset{\mu_L,\mu_R,\sigma_L,\sigma_R}{minimize} \sum_{i=1}^{N}(p_i y_i+n_i (1-y_i))+\frac{1-\xi}{N}\sum_{i=1}^{N} r_i, \mbox{such that}
\end{equation}
\begin{equation}
\label{MIP_9}
    \mu_R+\mathcal{MA}_{R_i}\geq\mu_i-0.5 \geq\mu_R -\mathcal{M}(1-\mathcal{A}_{R_i})+\epsilon, \forall i
\end{equation}
\begin{equation}
\label{MIP_10}
    \mathcal{M}(1-\mathcal{A}_{L_i})-\mu_L - \epsilon\geq\mu_i - 0.5 \geq-\mu_L- \mathcal{MA}_{L_i}, \forall i
\end{equation}
\begin{equation}
\label{MIP_11}
    \sigma_L+\mathcal{M}(1-\mathcal{A}_{DL_i})- \epsilon\geq\sigma_i\geq\sigma_L- \mathcal{MA}_{DL_i}, \forall i
\end{equation}
\begin{equation}
\label{MIP_12}
    \sigma_R+\mathcal{M}(1-\mathcal{A}_{DR_i})- \epsilon\geq \sigma_i\geq\sigma_R- \mathcal{MA}_{DR_i}, \forall i
\end{equation}
\begin{equation}
\label{MIP_13}
\mathcal{A}_{DL_i}+ \mathcal{A}_{L_i}\geq2p_i\geq\mathcal{A}_{DL_i}+ \mathcal{A}_{L_i}-1, \forall i
\end{equation}
\begin{equation}
\label{MIP_14}
\mathcal{A}_{DR_i}+ \mathcal{A}_{R_i}\geq2n_i\geq\mathcal{A}_{DR_i}+ \mathcal{A}_{R_i}-1, \forall i
\end{equation}
\begin{equation}
\label{MIP_15}
    p_i+n_i+r_i=1 \forall i
\end{equation}
\begin{equation}
\label{MIP_16}
    (\sum_{i=1}^{N}r_i)\leq r_{Cap}
\end{equation}
\begin{equation}
\label{MIP_17}
    \forall p_i,n_i,r_i,\mathcal{A}_{L_i},\mathcal{A}_{R_i},\mathcal{A}_{DL_i},\mathcal{A}_{DR_i}\in\{0,1\}
\end{equation}
\begin{equation}
\label{MIP_18}
    \mu_l,\mu_R,\sigma_L,\sigma_R,\xi\in\mathbb{R}
\end{equation}

We have assumed the cost of rejection to be constant for all samples. For cost-sensitive analysis, we refer the interested readers to \cite{RN146}.


\section{Discussion}

In this section, we first provide a numerical study towards validity of the rejection class. Thereafter, the main applications are discussed. Last but not least, open research gaps and future research scopes are listed at the end of this section. 
\subsection{Numerical Study Towards Validity of the Rejection Class}
\label{sec:numerical_study}
We produce results using three novel loss functions. VGG-16 \cite{10.48550/arxiv.1409.1556} and Resnet-18 \cite{10.48550/arxiv.1512.03385} are utilized as the training models, and the CIFAR-10 \cite{krizhevsky2009learning} and the SVHN \cite{37648} are selected as the experiment data. Different loss functions utilize different strategies for optimal solutions. To facilitate fare comparison, we followed the same training strategy for all the methods. During the first 50 epochs, we trained the models using the cross-entropy loss functions. Models are trained using the stochastic gradient descent method with a momentum of 0.9 and weight decay of 0.0005. At the beginning, 0.1 is used as the learning rate and then halved at $[10, 20, 25, 30, 35, 40, 45]^{th}$ epochs. The novel loss functions were used for the subsequent 10 epochs. Here, 0.001 is used for the learning rate.  

The selected loss functions use an extra output class for the rejection category. To make a comparison among different methods, we follow a fixed coverage strategy where scores are sorted out in an ascending or descending order, and then, first [98,95,90] percent of test samples are chosen for final accuracy calculation. For the NWLoss functions, we experimented with different values of $\alpha$ [0.5, 1.0, 1.5, 2.0, 4.0, 6.0]. For deep gambler functions, we use reward $o$ [2.0, 4.0, 6.0]. Though the results vary slightly for different hyperparameters, the overall conclusion remains the same. Here, we have only reported results for 2.0 for both $\alpha$ and $o$. 

All methods seem to perform similarly. The loss functions are expected to learn the complicated samples to reject through the extra class (rejection class) output. The other metrics that are based on two or more softmax outputs also produce similar results. Among other metrics, GI and $J_P$ almost always generate better results. Results in all the tables idicate that the extra class trained by those novel loss functions are capable of differentiating between easy and hard samples like those complex metrics, as the accuracy achieved by extra class score is very close to the accuracies achieved by other metrics. One might object that the other metrics always provide slightly better results than the extra class scores. We argue that all the scores are derived by training the models using those novel loss functions, not normal cross-entropy. Therefore, their better performances are ultimately credited towards those loss functions.

\begin{table}[!t]
\scriptsize
\caption{\label{tab:resnet18_cifar}Accuracy for the CIFAR-10 dataset using Resnet-18 model. Note, GI: (\ref{Gini_index}), Entropy: (\ref{predictive_uncertainty}) with M=1, LDAM: (\ref{LDAM}), $J_P$: (\ref{Uncertainty_metric}), Confidence: (\ref{confidence_credibility}), Relative difference: Relative difference between top two softmax output.}
\begin{filecontents*}{data/resnet18_cifar.csv}
,98,95,90,98,95,90,98,95,90
Extra class,90.72$\pm$0.18,91.79$\pm$0.19,93.31$\pm$0.17,90.61$\pm$0.22,91.68$\pm$0.24,93.23$\pm$0.22,90.64$\pm$0.17,91.71$\pm$0.19,93.21$\pm$0.22
GI,90.92$\pm$0.16,92.31$\pm$0.20,94.25$\pm$0.18,90.80$\pm$0.26,92.20$\pm$0.25,94.24$\pm$0.19,90.82$\pm$0.19,92.20$\pm$0.21,94.15$\pm$0.22
Entropy,90.90$\pm$0.19,92.24$\pm$0.17,94.18$\pm$0.18,90.77$\pm$0.24,92.16$\pm$0.24,94.16$\pm$0.21,90.76$\pm$0.18,92.12$\pm$0.20,94.03$\pm$0.18
LDAM,90.87$\pm$0.20,92.19$\pm$0.20,94.19$\pm$0.20,90.71$\pm$0.19,92.12$\pm$0.20,94.15$\pm$0.19,90.74$\pm$0.15,92.09$\pm$0.18,94.11$\pm$0.20
$J_P$,90.94$\pm$0.17,92.29$\pm$0.19,94.25$\pm$0.17,90.83$\pm$0.24,92.20$\pm$0.22,94.24$\pm$0.19,90.83$\pm$0.18,92.16$\pm$0.20,94.17$\pm$0.22
Confidence,90.67$\pm$0.21,91.93$\pm$0.22,94.01$\pm$0.20,90.53$\pm$0.20,91.87$\pm$0.16,94.03$\pm$0.24,90.55$\pm$0.18,91.85$\pm$0.20,93.93$\pm$0.17
Relative difference,90.82$\pm$0.23,92.16$\pm$0.21,94.15$\pm$0.21,90.67$\pm$0.19,92.06$\pm$0.20,94.13$\pm$0.23,90.70$\pm$0.17,92.05$\pm$0.18,94.09$\pm$0.20
\end{filecontents*}
\csvreader[
  tabular={|c|c|c|c||c|c|c||c|c|c|},
  table head=\hline\multirow{2}{*}{Methods} &
      \multicolumn{3}{c||}{DAC} &
      \multicolumn{3}{c||}{NWLoss} &
      \multicolumn{3}{c|}{Gambler}\\\cline{2-10}
      &98&95&90&98&95&90&98&95&90\\\hline,
  late after line=\\\hline
]{data/resnet18_cifar.csv}{}{
\csvcoli & \csvcolii & \csvcoliii & \csvcoliv & \csvcolv & \csvcolvi & \csvcolvii & \csvcolviii & \csvcolix & \csvcolx }

\end{table}
\begin{table}[!t]
\scriptsize
\caption{\label{tab:resnet18_svhn}Accuracy for the SVHN dataset using Resnet-18 model. Note, GI: (\ref{Gini_index}), Entropy: (\ref{predictive_uncertainty}) with M=1, LDAM: (\ref{LDAM}), $J_P$: (\ref{Uncertainty_metric}), Confidence: (\ref{confidence_credibility}), Relative difference: Relative difference between top two softmax output.}
\begin{filecontents*}{data/resnet18_svhn.csv}
,98,95,90,98,95,90,98,95,90
Extra class,96.860$\pm$0.094,97.624$\pm$0.089,98.376$\pm$0.060,96.867$\pm$0.158,97.617$\pm$0.132,98.353$\pm$0.072,96.864$\pm$0.112,97.638$\pm$0.072,98.379$\pm$0.064
GI,97.042$\pm$0.097,98.118$\pm$0.070,98.956$\pm$0.073,97.063$\pm$0.137,98.111$\pm$0.100,98.947$\pm$0.063,97.026$\pm$0.115,98.087$\pm$0.088,98.930$\pm$0.060
Entropy,96.996$\pm$0.093,98.041$\pm$0.077,98.926$\pm$0.077,97.024$\pm$0.146,98.036$\pm$0.121,98.933$\pm$0.060,96.991$\pm$0.114,98.006$\pm$0.084,98.902$\pm$0.064
LDAM,97.048$\pm$0.086,98.128$\pm$0.084,98.968$\pm$0.076,97.067$\pm$0.137,98.114$\pm$0.086,98.956$\pm$0.076,97.021$\pm$0.112,98.077$\pm$0.112,98.939$\pm$0.069
$J_P$,97.080$\pm$0.074,98.124$\pm$0.069,98.956$\pm$0.072,97.068$\pm$0.130,98.110$\pm$0.101,98.946$\pm$0.064,97.039$\pm$0.110,98.085$\pm$0.089,98.931$\pm$0.060
Confidence,96.936$\pm$0.075,98.085$\pm$0.084,98.958$\pm$0.074,96.950$\pm$0.131,98.085$\pm$0.091,98.949$\pm$0.077,96.899$\pm$0.105,98.033$\pm$0.114,98.935$\pm$0.071
Relative difference,97.034$\pm$0.081,98.122$\pm$0.082,98.962$\pm$0.076,97.052$\pm$0.139,98.106$\pm$0.086,98.952$\pm$0.077,97.000$\pm$0.108,98.070$\pm$0.114,98.938$\pm$0.068
\end{filecontents*}
\csvreader[
  tabular={|c|c|c|c||c|c|c||c|c|c|},
  table head=\hline\multirow{2}{*}{Methods} &
      \multicolumn{3}{c||}{DAC} &
      \multicolumn{3}{c||}{NWLoss} &
      \multicolumn{3}{c|}{Gambler}\\\cline{2-10}
      &98&95&90&98&95&90&98&95&90\\\hline,
  late after line=\\\hline
]{data/resnet18_svhn.csv}{}{%
\csvcoli & \csvcolii& \csvcoliii& \csvcoliv& \csvcolv& \csvcolvi& \csvcolvii& \csvcolviii& \csvcolix& \csvcolx}

\end{table}
\begin{table}[!t]
\scriptsize
\caption{\label{tab:vgg16_cifar}Accuracy for the CIFAR-10 dataset using VGG-16 model. Note, GI: (\ref{Gini_index}), Entropy: (\ref{predictive_uncertainty}) with M=1, LDAM: (\ref{LDAM}), $J_P$: (\ref{Uncertainty_metric}), Confidence: (\ref{confidence_credibility}), Relative difference: Relative difference between top two softmax output.}
\begin{filecontents*}{data/vgg16_cifar.csv}
,98,95,90,98,95,90,98,95,90
Extra class,92.86$\pm$0.75,94.09$\pm$0.66,95.66$\pm$0.54,93.01$\pm$0.28,94.15$\pm$0.30,95.75$\pm$0.25,92.97$\pm$0.31,94.14$\pm$0.38,95.72$\pm$0.36
GI,92.96$\pm$0.76,94.25$\pm$0.68,96.03$\pm$0.58,93.10$\pm$0.31,94.40$\pm$0.26,96.17$\pm$0.20,93.06$\pm$0.36,94.39$\pm$0.34,96.20$\pm$0.25
Entropy,92.98$\pm$0.75,94.26$\pm$0.66,96.03$\pm$0.58,93.12$\pm$0.32,94.41$\pm$0.25,96.18$\pm$0.21,93.06$\pm$0.35,94.41$\pm$0.35,96.19$\pm$0.28
LDAM,92.89$\pm$0.77,94.21$\pm$0.73,96.01$\pm$0.58,93.01$\pm$0.29,94.36$\pm$0.26,96.14$\pm$0.17,92.99$\pm$0.36,94.35$\pm$0.34,96.17$\pm$0.27
$J_P$,92.93$\pm$0.76,94.26$\pm$0.69,96.04$\pm$0.58,93.05$\pm$0.31,94.39$\pm$0.26,96.17$\pm$0.20,93.02$\pm$0.37,94.39$\pm$0.33,96.21$\pm$0.26
Confidence,92.81$\pm$0.81,94.18$\pm$0.75,96.00$\pm$0.57,92.94$\pm$0.29,94.35$\pm$0.28,96.12$\pm$0.18,92.94$\pm$0.36,94.32$\pm$0.33,96.16$\pm$0.27
Relative difference,92.87$\pm$0.78,94.21$\pm$0.74,96.01$\pm$0.58,93.00$\pm$0.29,94.36$\pm$0.26,96.12$\pm$0.18,92.97$\pm$0.37,94.35$\pm$0.34,96.16$\pm$0.27
\end{filecontents*}
\csvreader[
  tabular={|c|c|c|c||c|c|c||c|c|c|},
  table head=\hline\multirow{2}{*}{Methods} &
      \multicolumn{3}{c||}{DAC} &
      \multicolumn{3}{c||}{NWLoss} &
      \multicolumn{3}{c|}{Gambler}\\\cline{2-10}
      &98&95&90&98&95&90&98&95&90\\\hline,
  late after line=\\\hline
]{data/vgg16_cifar.csv}{}{%
\csvcoli & \csvcolii& \csvcoliii& \csvcoliv& \csvcolv& \csvcolvi& \csvcolvii& \csvcolviii& \csvcolix& \csvcolx}

\end{table}
\begin{table}[!t]
\scriptsize
\caption{\label{tab:vgg16_svhn}Accuracy for the SVHN dataset using VGG-16 model. Note, GI: (\ref{Gini_index}), Entropy: (\ref{predictive_uncertainty}) with M=1, LDAM: (\ref{LDAM}), $J_P$: (\ref{Uncertainty_metric}), Confidence: (\ref{confidence_credibility}), Relative difference: Relative difference between top two softmax outputs.}
\begin{filecontents*}{data/vgg16_svhn.csv}
,98,95,90,98,95,90,98,95,90
Extra class,97.125$\pm$0.104,97.918$\pm$0.083,98.487$\pm$0.063,97.235$\pm$0.115,98.053$\pm$0.092,98.594$\pm$0.060,97.204$\pm$0.103,98.020$\pm$0.028,98.598$\pm$0.053
GI,97.201$\pm$0.105,98.139$\pm$0.087,98.761$\pm$0.073,97.304$\pm$0.111,98.264$\pm$0.089,98.854$\pm$0.073,97.257$\pm$0.101,98.234$\pm$0.046,98.822$\pm$0.052
Entropy,97.200$\pm$0.109,98.131$\pm$0.090,98.744$\pm$0.069,97.307$\pm$0.113,98.261$\pm$0.094,98.839$\pm$0.067,97.259$\pm$0.110,98.217$\pm$0.048,98.810$\pm$0.053
LDAM,97.165$\pm$0.101,98.141$\pm$0.085,98.816$\pm$0.069,97.278$\pm$0.115,98.266$\pm$0.100,98.904$\pm$0.081,97.241$\pm$0.115,98.241$\pm$0.055,98.873$\pm$0.061
$J_P$,97.203$\pm$0.100,98.139$\pm$0.090,98.761$\pm$0.073,97.302$\pm$0.111,98.263$\pm$0.089,98.854$\pm$0.073,97.258$\pm$0.101,98.234$\pm$0.046,98.823$\pm$0.052
Confidence,97.146$\pm$0.113,98.142$\pm$0.086,98.816$\pm$0.069,97.258$\pm$0.118,98.257$\pm$0.102,98.898$\pm$0.080,97.229$\pm$0.118,98.239$\pm$0.056,98.873$\pm$0.061
Relative difference,97.165$\pm$0.105,98.143$\pm$0.086,98.815$\pm$0.068,97.275$\pm$0.117,98.260$\pm$0.101,98.898$\pm$0.079,97.240$\pm$0.116,98.241$\pm$0.055,98.873$\pm$0.061
\end{filecontents*}
\csvreader[
  tabular={|c|c|c|c||c|c|c||c|c|c|},
  table head=\hline\multirow{2}{*}{Methods} &
      \multicolumn{3}{c||}{DAC} &
      \multicolumn{3}{c||}{NWLoss} &
      \multicolumn{3}{c|}{Gambler}\\\cline{2-10}
      &98&95&90&98&95&90&98&95&90\\\hline,
  late after line=\\\hline
]{data/vgg16_svhn.csv}{}{%
\csvcoli & \csvcolii& \csvcoliii& \csvcoliv& \csvcolv& \csvcolvi& \csvcolvii& \csvcolviii& \csvcolix& \csvcolx}

\end{table}

\subsection{Applications}
\label{sec:application}
The main utility in the extra calculation accompanied by the reject option is observed when it is applied in the safety-critical application such as health care and autonomous driving. The reviewed works in this study encompass diversified application domains such as signature recognition, health care, astronomy, character recognition, digit recognition, bank note recognition, aerial feature recognition, video (audio+image) processing, human posture and activities recognition, object detection, robot localization in topological maps, building recognition, land-cover classification, fault diagnosis, speech recognition, road sign recognition, text classification, radio signal classification, hand gesture classification, TLS service classification, driving scene segmentation, climate prediction, tropical cyclone intensity, apparent age estimation, trading, novel environment recognition and visual question answering.

\begin{figure}[!t]
\centering
\subfloat[]{\includegraphics[scale=0.45]{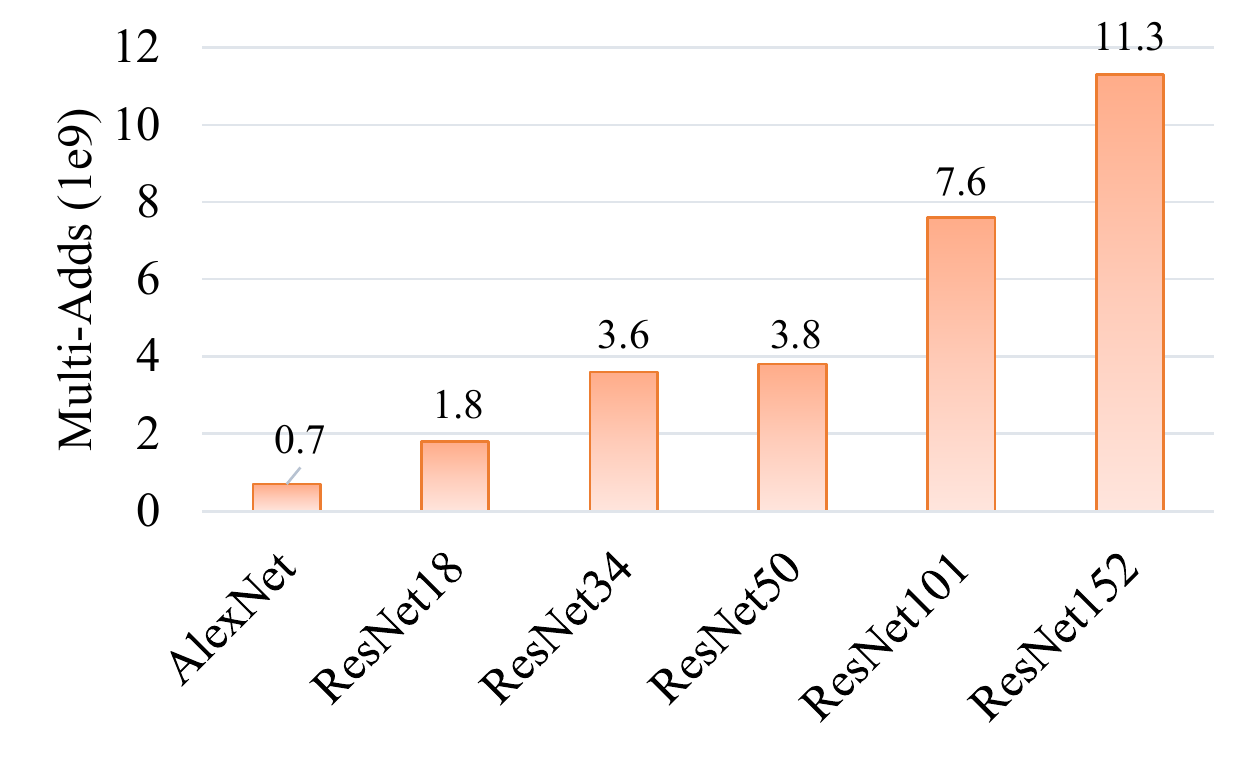}\label{imgnet_stats_1}}
\hfil
\subfloat[]{\includegraphics[scale=0.45]{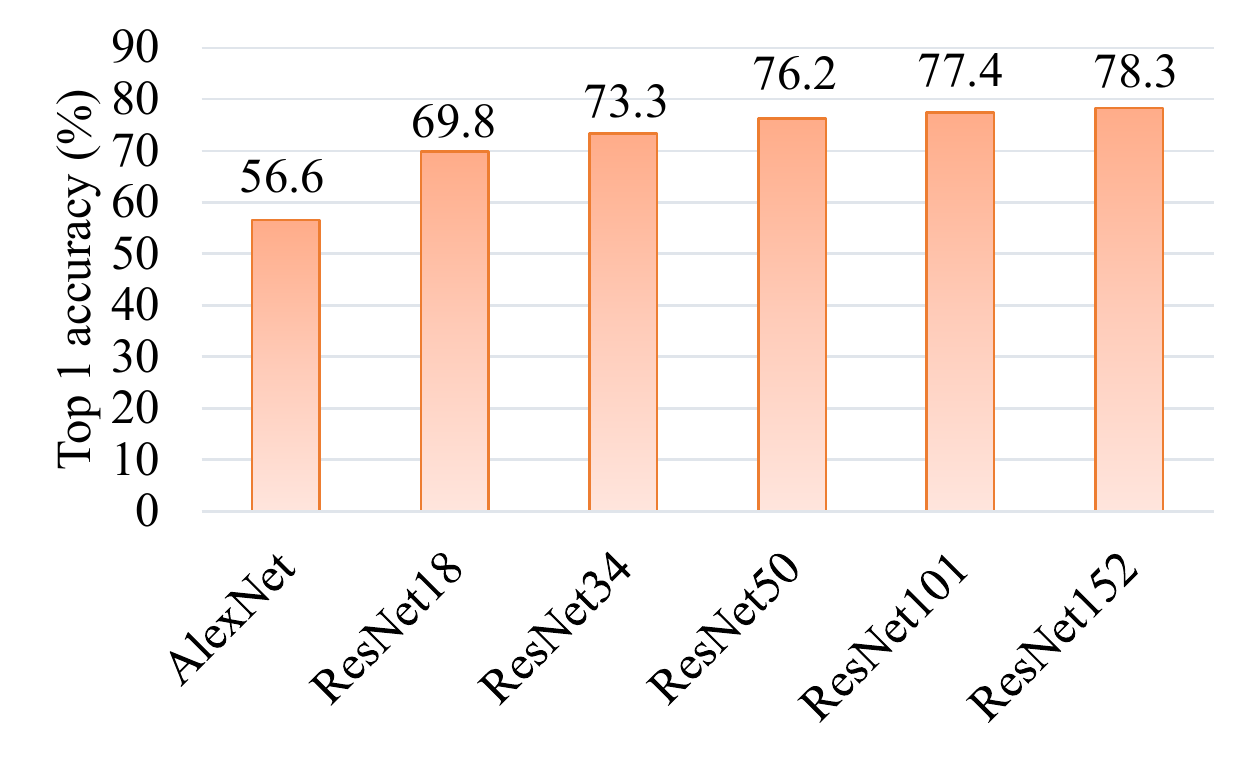}\label{imgnet_stats_2}}
\hfil
\subfloat[]{\includegraphics[scale=0.45]{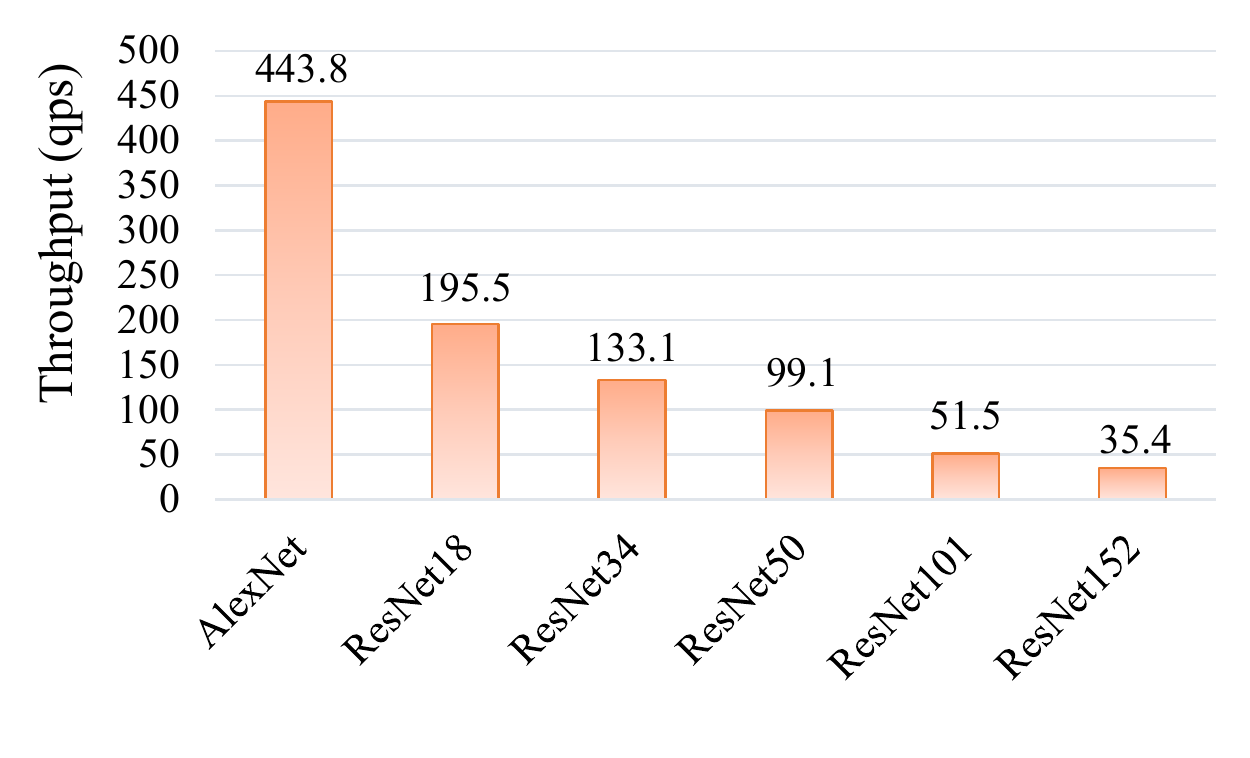}\label{imgnet_stats_3}}
\hfil
\subfloat[]{\includegraphics[scale=0.45]{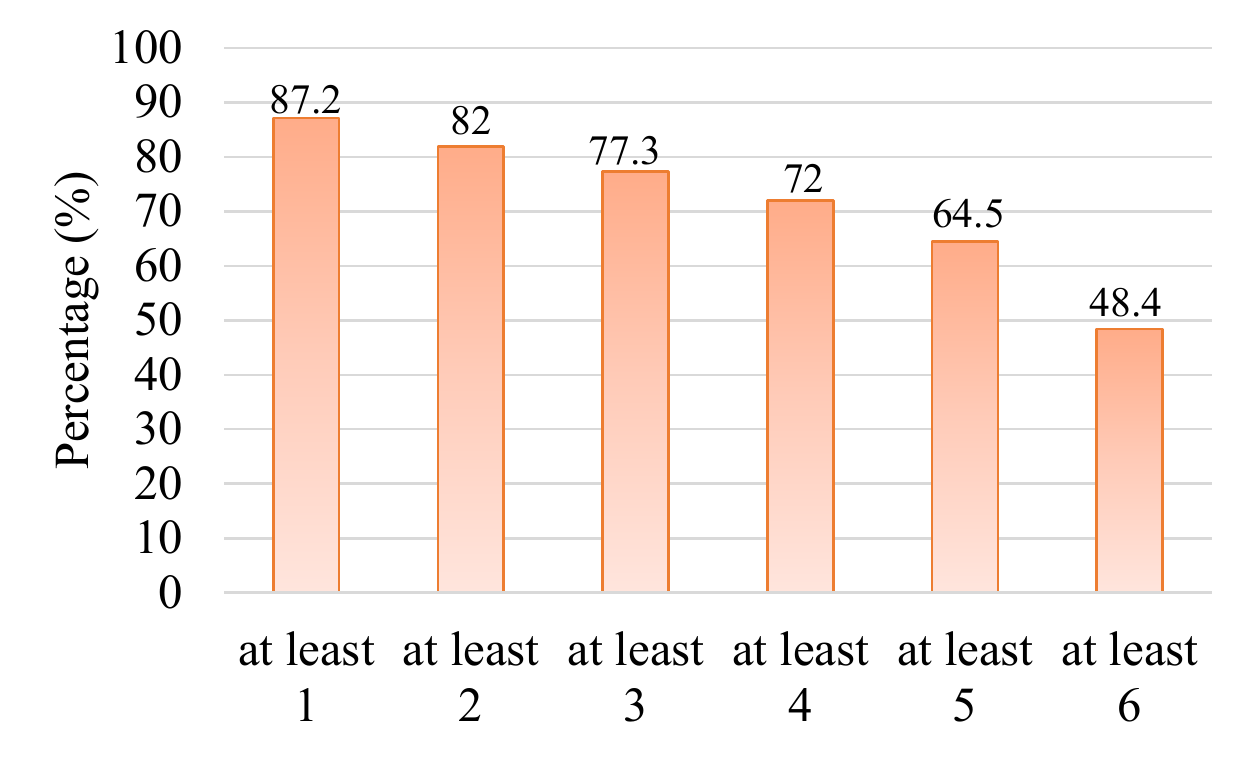}\label{imgnet_stats_4}}

\caption{ImageNet Model Statistics: (a) Number of Multiplication-Additions operation of the top ImageNet models. (b) The top-one prediction accuracy. (c) Throughputs (query per second) with batch size = 1. (d) The fraction
of images accurately classified by at least $K \in \{1 . . . 6\}$ of the benchmark models (reproduced from \cite{Wang2018IDKCF})}
\label{imagenet_stat}
\end{figure}

\begin{figure}[!t]
\centering
\includegraphics[scale=0.50]{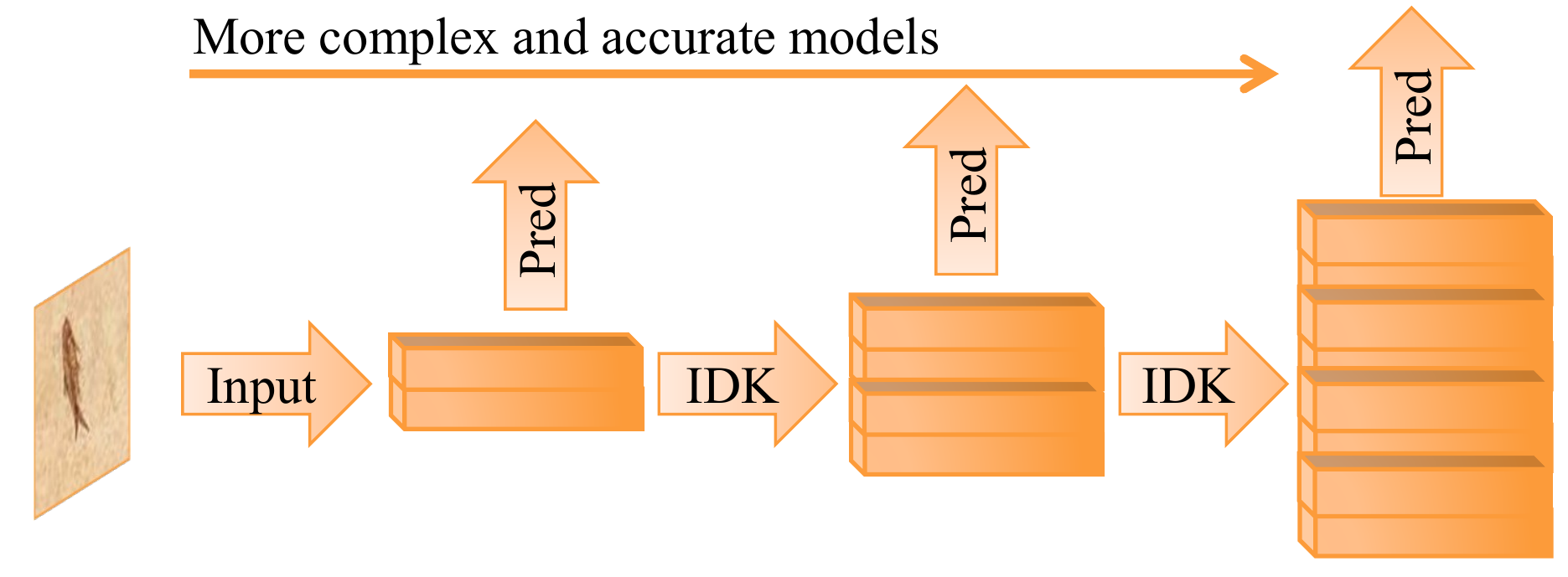}
\caption{ The schematic view of IDK cascade (reproduced from \cite{Wang2018IDKCF})}
\label{idk_cascade}
\end{figure}

Another very critical dimension that can utilize the prediction with reject option is where the response time of a DL model needs to be very low in order to avoid major operational failure of the overall system such as autonomous driving. Before diving to how rejection mechanism helps with the faster response of the DL model, we therefore concentrate on the statistics of the ImageNet dataset as indicated in Fig. \ref{imagenet_stat}.
 It is obvious from Figure \ref{imagenet_stat} that the increase in accuracy for the ImageNet dataset is much lower than the decrease in the throughputs of the benchmark models. At the same time, it is interesting to notice that nearly half of the inputs can be correctly classified by the very basic AlexNet model. Hence, without always using the more complex model, we can utilize the very light model to predict the easy samples and with the help of the rejection head, we can pass the harder samples to more advanced models in a cascaded manner. The general structure of the cascaded models named "IDK cascade" is shown in Fig. \ref{idk_cascade}.
Now, given multiple classifiers with a reject option, how to cascade them in a systematic way so that the average prediction time is minimized under different constraints is beyond the scope of this study. We refer the interested readers to \cite{10.1145/3453417.3453425,Baruah2022} for a detailed analysis of the IDK cascade.

\subsection{Future Research Scopes}
\label{sec:future_scope}
As we already stated, there have been limited works in this particular area in the context of the NNs. The main areas that need further investigation can be discussed under the following categories:
\begin{itemize}
    \item Data Collection and Scoring: All of the available datasets are mainly collected and labeled focusing on the prediction accuracy of the related task. For example, `mixup' augmentation is used in \cite{10.1007/978-3-030-86340-1_23} to come up with a novel difficulty score for the rejection cost.  Thus, further investigation is required to develop a data collection protocol and the data scoring protocol (such as focusing on the difficulty of inference of the data) so that those scores can be directly used in the rejection strategy.
    \item Explainable AI (XAI) \cite{gunning2017explainable}: Explainability is one of the key factors for the adoption of ML models in many mission-critical applications \cite{https://doi.org/10.1002/widm.1391}. XAI is the domain in machine learning research focusing on  the intelligent reasons for the decision of an ML model. Now, Interpretability is one of the most discussed topics in the field of XAI. Often, ‘interpretability’ and ‘explainability’ are used synonymously in literature \cite{8466590}. According to \cite{biran2017explanation}, “Explanation is considered closely related to the concept of interpretability”. Regarding interpretability, we can follow the definition given by \cite{murdoch2019interpretable}, “We define interpretable machine learning as the use of machine-learning models for the extraction of relevant knowledge about domain relationships contained in data...”. Usually, the more accurate models (such as DNNs) are more difficult to understand and interpret \cite{doi:10.1126/scirobotics.aay7120}. Following \cite{DBLP:journals/corr/abs-2012-01805}, some of the goals that can be achieved with interpretability are trust, causality, fairness, reliability, and privacy as depicted in Fig. \ref{interpretability} . Here, reliability indicates the resistance of ML systems to noisy inputs and reasonable domain shift. The reject option also tries to address uncertain inputs due to noise and/or domain shift. We believe that the reject option can provide much in the field of XAI in terms of reliability.

\begin{figure}[!t]
\centering
\includegraphics[scale=0.36]{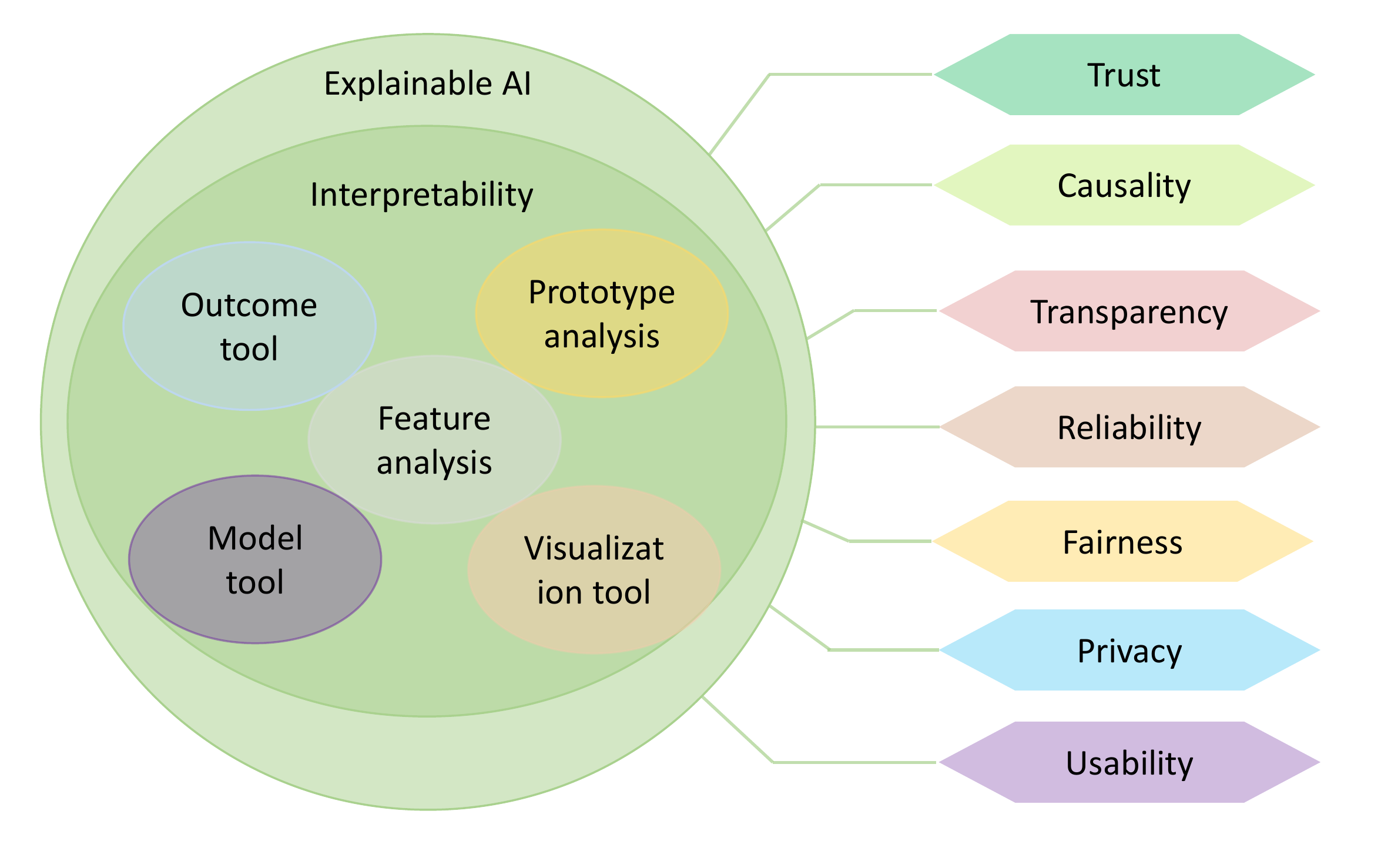}
\caption{Major goals of explainable AI. It should be noted that reliability can be achieved with the reject option.}
\label{interpretability}
\end{figure} 
    
    \item Network Architecture: Although the authors in \cite{RN381} utilized the features from the hidden layers of a trained CNN, they trained other NNs using these features to make the final \textit{rejection} decision. One of the future works may be to develop strategies or dedicated layers to directly obtain the decision metrics from the intermediate layers so that it becomes readily available after the training of the NNs is complete. We believe that works done in \cite{10.48550/arxiv.2102.11582,Papernot2018DeepKN} can serve as a motivational guideline to this direction.
    \item Cost Function: All the cost functions that include the rejection mechanism into them usually come with an increased number of hyper-parameters. Sometimes, to determine suitable values for these hyper-parameters, another optimization strategy is needed which increases the computation cost. Even when no extra hyper-parameter is introduced, extra training data related to the rejection class are needed \cite{Thulasidasan2021AnEB}. An interesting direction of research may be to develop an objective function with a minimum number of hyper-parameters while avoiding the use of rejection class data at the same time.
    \item Robustness: While DNNs have achieved outstanding performance in a wide range of settings, they are still susceptible to small perturbations in inputs resulting in high variation in embedding space. Existing literature focus on improving robustness (local \cite{ DBLP:journals/corr/abs-2010-04821} and/or global \cite{ ijcai2019p824}) against these perturbations. Some of the popular approaches include gradient diversity regularization \cite{ DBLP:journals/corr/abs-2107-02425}, $L_2$ or $L_\infty$ Normalization \cite{ Yu2022} and stability training \cite{ DBLP:journals/corr/ZhengSLG16} etc. We noticed that there has been little effort to include this kind of adversarial robustness into reject option.
    \item Logical Reasoning: Recently, researchers have been developing increasing interest in blending logical or symbolic reasoning into DNN architecture \cite{giunchiglia2022deep}. For example, authors in \cite{DBLP:journals/corr/SerafiniG16} propose the logic tensor network to automatically integrate reasoning and learning . A differentiable (smoothed) maximum satisfiability (MAXSAT) solver is developed that can be used in the loop of the end-to-end learning system \cite{DBLP:journals/corr/abs-1905-12149}. A comprehensive analysis of the deep learning approaches that utilize logical constraint in the first-order logic is done in \cite{giunchiglia2022deep}. However, we notice that there has been no attempt to integrate the reject option with logical reasoning, which might be an interesting future research direction.
\end{itemize}

\section{Conclusion}
\label{sec:conclusion}
With the digitalization of data in almost every sector of modern society, automated decision-making is becoming more and more prevalent, and machine learning models (especially DNNs) are playing a pivotal role. But, the application of the DNN models in safety-critical applications is still not approved by many experts due to their lack of knowledge-awareness. Introducing the knowledge domain into the ML model while retaining \textit{state-of-the-art} prediction accuracy is of the utmost necessity for real-world deployment. Hence, we have focused on a sub-field of ML strategy, namely prediction with rejection (or abstention or selection), in the context of neural networks. We classified the available rejection strategies from reviewed articles, discussed novel cost functions blending prediction and rejection functions, and discussed scoring methods for rejection and different threshold determination methods to clarify the boundary between prediction and rejection regions. Empirical studies are done to justify the current trend of utilizing an extra prediction category denoting the knowledge awareness of the ML models (uncertainty in prediction). Finally, we have outlined important research gaps for future investigations.
\begin{acks}
We thank Dr. Diederik P. Kingma from Google Brain, USA, for checking the article and providing valuable feedback and comments. This research was partially supported by the Australian Research Council’s Discovery Projects funding scheme (project DP190102181 and DP210101465).
\end{acks}

\bibliographystyle{ACM-Reference-Format}
\bibliography{index}

\appendix

\end{document}